\documentclass[sigconf, most]{acmart}

\setcopyright{none}               
\renewcommand\footnotetextcopyrightpermission[1]{} 
\AtBeginDocument{%
  }

\setcopyright{none}      
\copyrightyear{2026}     
\acmYear{2026}
\acmDOI{}                
\acmISBN{}               

\acmConference[Preprint]{Under Review}{March 2026}{Durham, NC}

\usepackage{algorithm}
\usepackage[noend]{algpseudocode}
\usepackage{bbm, bm}
\usepackage{booktabs} 
\usepackage{graphicx} 
\usepackage{todonotes}
\usepackage{xcolor}
\usepackage{booktabs}
\usepackage{tabularx}
\usepackage{multirow}
\usepackage{makecell}

\usepackage{tcolorbox}
\usepackage{booktabs}
\usepackage{xcolor}

\usepackage{array}
\usepackage{colortbl}
\usepackage{xcolor}

\usepackage{amsthm}
\theoremstyle{definition}
\newtheorem{example}{Example}[section]

\usepackage{tcolorbox}

\newcounter{myboxcount}[section]
\renewcommand{\themyboxcount}{\thesection.\arabic{myboxcount}}

\ifdefined\promptbox\else
\newtcolorbox{promptbox}[2][]{%
    code={\refstepcounter{myboxcount}}, 
    width=\textwidth,
    colback=white,
    colframe=black!70,
    sharp corners,
    boxrule=0.8pt,
    fonttitle=\small\bfseries,
    coltitle=white,
    title={Box \themyboxcount: #2}, 
    #1
}
\fi

\NewDocumentCommand{\matt}
{ mO{} }{\textcolor{teal}{\textsuperscript{\textit{Matt}}\textsf{\textbf{\small[#1]}}}}

\newcommand{\nop}[1]{}

\begin{document}
\begin{sloppy}
\title{\textsc{MPCEval}: A Benchmark for Multi-Party Conversation Generation}











\author{Minxing Zhang\textsuperscript{1, \dag} \hspace{5mm} Yi Yang\textsuperscript{1, \dag} \hspace{5mm} Zhuofan Jia\textsuperscript{1} \hspace{5mm} Xuan Yang\textsuperscript{1} \hspace{5mm} Jian Pei\textsuperscript{1}\\ Yuchen Zang\textsuperscript{2} \hspace{1mm} Xingwang Deng\textsuperscript{2} \hspace{1mm} Xianglong Chen\textsuperscript{2}}
\thanks{\dag Both authors contributed equally to this research.}
\affiliation{%
  \institution{\textsuperscript{1}Duke University \hspace{15mm} \textsuperscript{2}Tanka AI}
  \country{USA}
}
\email{{minxing.zhang, owen.yang, zhuofan.jia, xuan.yang, j.pei}@duke.edu}
\email{{zangyuchen, dengxingwang, xianglong.chen}@tanka.ai}

\renewcommand{\shortauthors}{Zhang et al.}

\begin{abstract}
Multi-party conversation generation, such as smart reply and collaborative assistants, is an increasingly important capability of generative AI, yet its evaluation remains a critical bottleneck. Compared to two-party dialogue, multi-party settings introduce distinct challenges, including complex turn-taking, role-dependent speaker behavior, long-range conversational structure, and multiple equally valid continuations. Accordingly, we introduce \textsc{MPCEval}, a task-aware evaluation and benchmarking suite for multi-party conversation generation. \textsc{MPCEval} decomposes generation quality into speaker modeling, content quality, and speaker--content consistency, and explicitly distinguishes local next-turn prediction from global full-conversation generation. It provides novel, quantitative, reference-free, and reproducible metrics that scale across datasets and models. We apply \textsc{MPCEval} to diverse public and real-world datasets and evaluate modern generation methods alongside human-authored conversations. The results reveal systematic, dimension-specific model characteristics in participation balance, content progression and novelty, and speaker--content consistency, demonstrating that evaluation objectives critically shape model assessment and that single-score evaluation obscures fundamental differences in multi-party conversational behavior. The implementation of \textsc{MPCEval} and the associated evaluation code are publicly available at \url{https://github.com/Owen-Yang-18/MPCEval}.
\end{abstract}

\maketitle







\section{Introduction}
\label{sec:introduction}

Multi-party conversation generation aims to synthesize coherent dialogues involving three or more participants~\cite{gu2022says}. This capability underpins collaborative AI systems such as virtual meeting assistants, group decision-support agents, and multi-user chatbots, where participants with distinct roles and expertise jointly pursue shared objectives~\cite{li2024chatmdg, penzo2025don, ganesh2023survey, chernyavskiy2023transformer, wei2023multi, zhu-etal-2022-multi}. Compared to two-party dialogue, multi-party interactions exhibit different dynamics, including complex turn-taking, implicit coordination, role-dependent behavior, and long-range conversational dependencies~\cite{ganesh2023survey,zhang2018addressee}. While these properties better reflect real-world communication, they also substantially complicate both generation and evaluation~\cite{ganesh2023survey, wei2023multi}.

Recent advances in large language models (LLMs)~\cite{10.1145/3794858} have driven rapid progress in multi-party conversation generation~\cite{addlesee2024multi, chernyavskiy2023transformer}. However, evaluation methodology has lagged behind. \textbf{Existing benchmarks and metrics provide limited guidance for comparing generation models or diagnosing their failures}, making it difficult to disentangle improvements in speaker modeling, content quality, or role consistency from superficial optimization of specific metrics. Consequently, models with fundamentally different conversational behaviors may receive similar scores, while models excelling in one dimension may be unfairly penalized in others, hindering both scientific understanding and practical deployment.

Evaluating multi-party conversation generation poses several challenges. \textbf{First, conversational quality is inherently multi-dimensional}, depending jointly on who speaks, what is said, and whether content aligns with the speaker's role and prior behavior; collapsing these aspects into a single score obscures critical trade-offs~\cite{liu2016not,ouchi2016addressee,zhang2018addressee}. \textbf{Second, evaluation is task-dependent}: next-message prediction emphasizes local plausibility, whereas full-conversation generation requires assessing global properties such as participation balance, information flow, and long-term role consistency. \textbf{Third, conversation is intrinsically open-ended}, as multiple continuations may be equally valid for a given context, rendering reference-based metrics fundamentally inadequate~\cite{zhang2024reviseval, zhang2019bertscore, yuan2021bartscore, kim2022generating, kim2021neuralwoz,shi2023rade}. \textbf{Finally, evaluation frameworks must be extensible} to accommodate emerging conversational dimensions and paradigms.

To address these challenges, we introduce \textbf{\textsc{MPCEval}}, a task-aware and decomposed evaluation framework for multi-party conversation generation. \textsc{MPCEval} explicitly separates speaker modeling, content quality, and speaker--content consistency, and aligns each dimension with the appropriate evaluation granularity. It distinguishes local evaluation for next-message prediction from global evaluation for entire conversation generation, ensuring that metrics reflect underlying conversational objectives rather than conflating distinct phenomena. Building on this structure, \textsc{MPCEval} provides a suite of novel, quantitative, reference-free, and reproducible metrics that scale across datasets, tasks, and models, enabling fine-grained and principled comparison of generation methods. \textbf{Experiments via \textsc{MPCEval} further demonstrate that human-authored conversations are not uniformly superior across all evaluation dimensions, and that different models excel in distinct aspects of multi-party dialogue generation, providing nuanced understanding towards each model's specific strengths and characteristics.}

\textbf{Existing evaluation paradigms systematically mislead model comparison} in the multi-party setting. Traditional sequence-level metrics such as BLEU, ROUGE, and perplexity~\cite{zhu2018texygen, lin2004rouge, kim2022generating, jelinek1977perplexity, wu2022dg2} operate at the sequence level and fail to capture long-range interaction dynamics. Single-reference-based metrics like BERTScore~\cite{zhang2019bertscore, gao2019interconnected} assume a unique correct response and misrepresent the diversity of valid continuations. Consequently, lexical-level similarity is often rewarded while ignoring fundamental qualities such as role consistency, meaningful content progression, and interaction structure. LLM-as-a-judge approaches~\cite{gu2024survey, liu2023g, soudani2024survey, zheng2023judging} avoid explicit references but suffer from limited reproducibility and reliability, and sensitivity to prompt design~\cite{ji2023survey}. 

In this paper, we make \textbf{four main contributions}. First, we introduce \textsc{MPCEval}, the first benchmark designed for evaluating multi-party conversation generation. Second, we propose a task-aware evaluation framework that distinguishes local next-message prediction from global full-conversation generation. Third, we develop a comprehensive set of quantitative, reference-free, and reproducible metrics that capture structural and semantic properties of multi-party dialogue beyond existing evaluations. Fourth, we design \textsc{MPCEval} as an open and extensible framework, enabling future evaluation dimensions to be integrated as the field evolves.

\textsc{MPCEval} is developed through a close collaboration between academia and industry, and is designed to support both principled benchmarking research and real-world deployment; it is being integrated into the evaluation workflow of Tanka AI and will be released as an open-source benchmark for public use.
\section{Related Work}
\label{sec:related_work}

Our work relates to prior research on conversational evaluation and multi-party conversation generation. In this section, we focus on limitations of existing methods in multi-party settings and how these gaps motivate \textsc{MPCEval}.

\subsection{Evaluation of Conversation Generation}

A broad range of evaluation methods has been developed for two-party dialogue, including \emph{reference-based metrics}, \emph{reference-free automatic metrics}, and \emph{human evaluation protocols}. While effective for narrow objectives, these approaches do not generalize well to open-ended, multi-party interaction.

Reference-based metrics compare generated responses against fixed human-authored texts, including BLEU and ROUGE~\cite{papineni2002bleu, lin2004rouge, jelinek1977perplexity, zhu2018texygen, kim2022generating, wu2022dg2} and embedding-based variants like BERTScore~\cite{zhang2019bertscore}. As discussed in Section~\ref{sec:introduction}, these metrics implicitly assume a unique correct response, an assumption that breaks down when multiple conversational continuations are equally valid. Consequently, they often penalize appropriate responses that diverge from the reference and fail to capture complex properties such as turn-taking, topical progression, or speaker attribution. Moreover, because these metrics operate primarily at the sequence level, they do not extend naturally to global evaluation of entire conversations.

To mitigate dependence on human-authored references, prior work has proposed reference-free metrics that assess intrinsic properties of generated text. Diversity-based measures such as Distinct-$n$~\cite{li2016diversity, zhang2018generating} quantify lexical variety but provide limited insight into coherence or contextual relevance. More recent LLM-as-a-judge approaches~\cite{gu2024survey, mehri2020usr, zhong2022towards, liu2023g} use large language models (LLMs) to score responses holistically, but suffer from limited reproducibility due to sensitivity to prompts, evaluator models, and hallucinated judgments. More broadly, existing automatic metrics remain narrow in scope~\cite{kim2021neuralwoz, yuan2021bartscore, zhong2022towards}: for example, BARTScore~\cite{yuan2021bartscore} reflects token-level likelihood, while G-Eval~\cite{liu2023g} emphasizes engagement, leaving many dimensions of conversational quality unmeasured.

Human evaluation is often considered the gold standard for assessing coherence, engagement, and naturalness~\cite{deriu2021survey, smith2022human}, but it is costly, and difficult to standardize, limiting its capability for systematic and scalable benchmarking. Overall, existing evaluation paradigms neither support fair and consistent comparison across models and tasks nor capture the multi-dimensional properties central to multi-party interaction. In contrast, \textsc{MPCEval} provides novel, reference-free, quantitative, reproducible, and comprehensive metrics that jointly assess speaker modeling, content quality, and speaker--content consistency at both local and global levels.

\subsection{Multi-Party Conversation Generation}

Research on multi-party conversation generation spans diverse objectives and methodologies. Existing approaches can be broadly categorized by task into next-message prediction~\cite{tan2023chatgpt, wei2023multi} and full-conversation generation~\cite{penzo2025don}, including both continuation from a given history and generation  under topic or participant constraints.

From a methodological perspective, contemporary approaches fall into prompt-based and training-based paradigms. Prompt-based methods leverage the zero-shot or few-shot capabilities of state-of-the-art (SOTA) LLMs~\cite{tan2023chatgpt, penzo2025don}, whereas training-based methods rely on supervised fine-tuning with annotated multi-party dialogue data, such as speaker labels or discourse structure~\cite{wei2023multi, mahajan2024persona, chernyavskiy2023transformer, li2024chatmdg, gu2022hetermpc}. Many training-based methods further assume access to explicit addressing information~\cite{chernyavskiy2023transformer, li2024chatmdg, gu2022hetermpc} or participant profile metadata~\cite{mahajan2024persona}, which are costly and often impractical to obtain at scale.

Despite substantial progress in generation models, the field lacks a dedicated evaluation benchmark tailored to multi-party interaction. Existing evaluation practices are largely inherited from two-party dialogue and do not adequately account for complex speaker dynamics, long-range interaction patterns, or multiple valid conversational trajectories. To the best of our knowledge, \textsc{MPCEval} is the first benchmark to offer a standardized, task-aware, and comprehensive evaluation framework for multi-party conversation generation, enabling principled comparison of diverse generation methods across both next-message prediction and full-conversation generation tasks, while remaining extensible to emerging evaluation dimensions as richer metadata becomes available. 

In short, existing work lacks an evaluation framework for jointly assessing task-aware and decomposed quality dimensions in multi-party conversations---a gap \textsc{MPCEval} is designed to fill.
\section{Objectives and Overall Design}
\label{sec:benchmark-objective}

This section outlines the design objectives and structure of \textsc{MPCEval}, explaining its design choices and how they address the challenges of evaluating multi-party conversation generation.

\subsection{What Does \textsc{MPCEval} Measure?}\label{sec:measure_objective}

\textsc{MPCEval} is designed around a minimal yet expressive decomposition of multi-party conversation quality into three complementary dimensions: \emph{speaker modeling}, \emph{content quality}, and \emph{speaker--content consistency}. This decomposition reflects three orthogonal questions that any conversational system must answer: who should speak next, what should be said, and whether the content is consistent with the attributed speaker. These dimensions jointly capture the core structural and semantic properties of multi-party interaction without collapsing them into a single opaque score.

Evaluation in \textsc{MPCEval} is explicitly task-aware. We distinguish two evaluation tasks that impose fundamentally different requirements on generation models. In \emph{next-message prediction} (local evaluation), a model is given a conversation history, optionally augmented with speaker profiles or explicit conversation objectives, and asked to generate the next turn. Evaluation in this setting emphasizes \emph{local plausibility}, including the contextual appropriateness of the predicted speaker and the coherence of the generated utterance with respect to the immediate dialogue state.

In contrast, \emph{full-conversation generation} (global evaluation) requires a model to produce a multi-turn dialogue, either by extending a given history or by generating a conversation from scratch under specified constraints such as topics or participant lists. Evaluation at this level emphasizes \emph{global plausibility}, capturing long-range properties including participation balance, information flow, sustained speaker roles, and task completion when conversation objectives are defined. The explicit separation between local and global evaluation prevents metrics designed for short-term plausibility from being misapplied to long-horizon conversational structure.

\subsection{Datasets}
\label{sec:dataset-method}

\textsc{MPCEval} is dataset-agnostic (for text-based conversations) and is designed to support evaluation across heterogeneous conversational settings with varying levels of annotation. When speaker labels are unavailable, \textsc{MPCEval} can still assess content generation quality; when speaker information is present, it additionally evaluates speaker modeling and speaker--content consistency. If richer metadata such as speaker profiles are provided, \textsc{MPCEval} can further incorporate this information to derive more fine-grained evaluation metrics. In this paper, we select three representative multi-party conversation datasets, covering different real-world scenarios: \textbf{DeliData}~\cite{karadzhov2023delidata}, featuring collaborative task-solving dialogues with explicit objectives; \textbf{MPDD}~\cite{chen2020mpdd}, derived from movie scripts with rich speaker interactions; and \textbf{Tanka}, a real-world enterprise communication dataset characterized by long, information-dense discussions. Together, these datasets span diverse domains, interaction lengths, and speaker configurations, enabling us to examine the generalizability of the proposed metrics in \textsc{MPCEval}. Detailed dataset descriptions are provided in Appendix~\ref{sec:dataset-detail}.

\subsection{Generation Methods}

A key design objective of \textsc{MPCEval} is method-agnosticism. The framework is designed to evaluate generation methods with fundamentally different modeling assumptions, including both prompt-based and training-based approaches. In our experiments, we benchmark a diverse set of state-of-the-art (SOTA) generation methods across both next-message prediction and full-conversation generation tasks. Given the prevalence of prompting-based approaches in practice, \textsc{MPCEval} emphasizes evaluation of the zero-shot capabilities of modern LLMs, while remaining compatible with training-based or other methods. This method-agnostic design ensures that evaluation reflects the distinct characteristics that each model prioritizes and captures in conversation generation, rather than artifacts of a particular modeling paradigm, and allows new generation methods to be incorporated without modifying the evaluation pipeline.

\begin{figure}[t]
    \centering
    \includegraphics[width=\columnwidth]{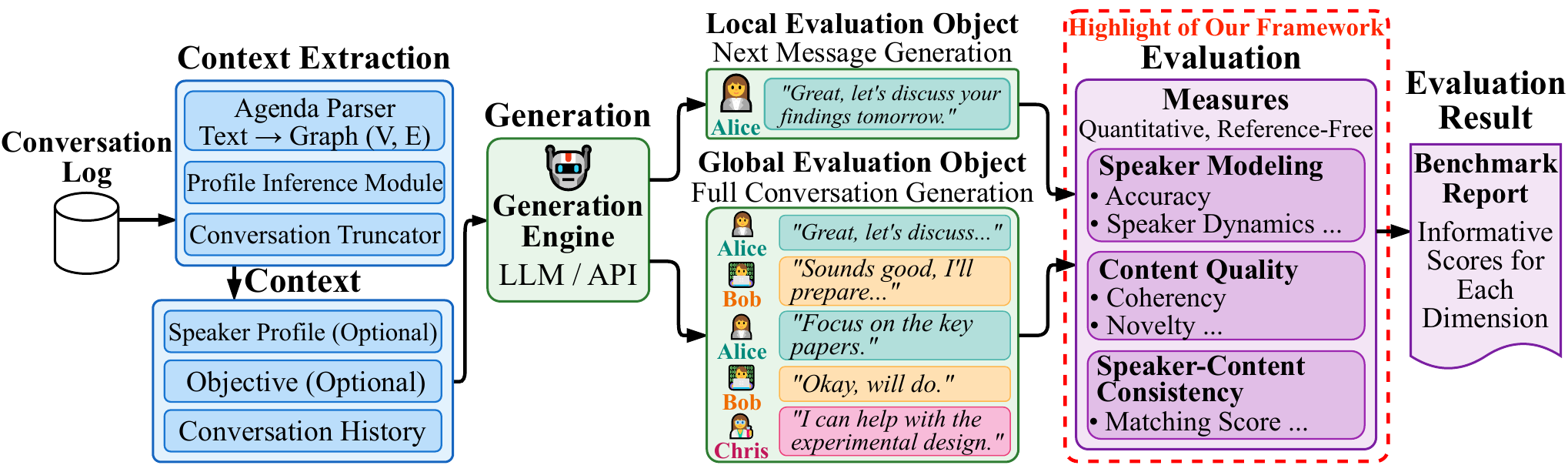}
    \caption{Overall design of the \textsc{MPCEval} framework.}
    \label{fig:global-measurement-design}
\end{figure}

\subsection{Evaluation Workflow and Overall Design}
\label{sec:global-measurement-design}

As shown in Figure~\ref{fig:global-measurement-design}, \textsc{MPCEval} is organized around three core components: \emph{context}, \emph{evaluation objects}, and \emph{measures}. This modular design supports both fine-grained and holistic evaluation of multi-party conversation generation, while remaining extensible to new evaluation dimensions.

The \emph{context} specifies the information available for generation and evaluation, including the conversation history and optional metadata such as speaker profiles or explicit objectives. This design allows \textsc{MPCEval} to operate robustly across diverse scenarios, enabling metadata-driven metrics when such information is available and principled reference-free evaluation otherwise.

Given a context, \textsc{MPCEval} evaluates generation objects at two levels of granularity. \emph{Local evaluation} assesses a single predicted next turn, focusing on speaker selection and content appropriateness given the immediate dialogue state. \emph{Global evaluation} assesses an entire generated continuation---either extending a provided history or generating from scratch---emphasizing long-range coherence, interaction dynamics, and overall conversational progression. Together, these evaluation objects capture short-term plausibility and long-horizon conversational structure.

For both local and global evaluation, \textsc{MPCEval} organizes measures along three complementary aspects: \emph{speaker modeling}, \emph{content quality}, and \emph{speaker--content consistency}. Rather than collapsing all aspects of quality into a single scalar score, the framework produces a multi-dimensional evaluation profile. This design choice is intentional: it enables the identification of trade-offs that are invisible to aggregate metrics. For example, one model may excel at speaker selection but exhibit weak content progression, while another may generate coherent content with imbalanced participation. By exposing these differences explicitly, \textsc{MPCEval} supports diagnostic analysis and principled model comparison.

Overall, this unified design enables evaluation capabilities that existing benchmarks lack, including task-aware assessment across local and global generation settings, explicit decomposition of conversational quality into interpretable dimensions, and reference-free, reproducible  measurement of multi-party interaction dynamics. We instantiate this design with a set of novel, quantitative, reference-free, reproducible, and comprehensive metrics for local evaluation (Section~\ref{sec:local-eval}) and global evaluation (Section~\ref{sec:global-eval}). Table~\ref{tab:mpceval_metrics} summarizes all proposed metrics, together with their scope, interpretation, and the conversational quality dimensions they measure.

\begin{table*}[t]
\centering
\caption{Summary of \textsc{MPCEval} metrics, which are quantitative, reference-free, and reproducible. $\uparrow$ indicates a higher value.}
\label{tab:mpceval_metrics}
\small
\setlength{\tabcolsep}{5pt}
\resizebox{0.85\textwidth}{!}{
\begin{tabularx}{\textwidth}{p{1.0cm} X}
\toprule
\textbf{Aspect} & \textbf{Metric and Description} \\
\midrule\hline

\multicolumn{2}{l}{\textbf{Local Measures}} \\
\hline
\multirow{5}{*}{Speaker}
& \textbf{DNR} (Direct Name Reference): binary indicator of whether the predicted next speaker is explicitly addressed in the recent context. \\
& \textbf{IR} (Implicit Reference): higher values indicate the predicted speaker participated more recently when no explicit addressing cue is present. \\
& \textbf{PF} (Participation Frequency): fraction of recent turns contributed by the speaker; a higher score means more frequent short-term engagement. \\
& \textbf{LS-ES} (Embedding Similarity): similarity between recent context and the predicted speaker's prior utterances. $\uparrow$ score means $\uparrow$ similarity. \\
& \textbf{LS-TA} (Topic Alignment): similarity between topic distributions of recent context and the speaker's prior utterances. $\uparrow$ score means $\uparrow$ similarity. \\

\midrule
\multirow{7}{*}{Content}
& \textbf{AP} (Agenda Progression): relevance and novelty of the response w.r.t. the current active or the next agenda item. $\uparrow$ score means $\uparrow$ progression.\\
& \textbf{LNR-E-w} (Weighted Emb.-aware LNR): IDF-weighted novelty rate of lexical units in the response. ($\approx$0.5: balance, 0: redundancy, 1: divergence) \\
& \textbf{M-SNS} (Message-level Semantic Novelty): emb. distance between response and recent turns. ($\approx$0.5: balance, 0: redundancy, 1: divergence) \\
& \textbf{DAF} (Dialogue-Act Transition Fit): likelihood that the response functionality (e.g., answer) follows context patterns. $\uparrow$ score means $\uparrow$ likelihood. \\
& \textbf{LL} (Log-Likelihood): normalized log likelihood assigned by an LLM to the response, conditioned on the context. \\
& \textbf{TES} (Topic Expansion Score): the response's expansion beyond the conversation's dominant topics. ($\approx$0.5: balance, 0: redundancy, 1: off-topic). \\

\midrule
\multirow{1}{*}{\makecell{Speaker--\\Content}}
& \textbf{LSCC-ES} (Embedding-Based Local Speaker--Content Consistency): embedding similarity between the response and the predicted speaker's utterances in the recent context, with optional background-augmented speaker representation for sparse histories. $\uparrow$ score means $\uparrow$ similarity. \\

\midrule
\multicolumn{2}{l}{\textbf{Global Measures}} \\
\midrule

\multirow{2}{*}{Speaker}
& \textbf{NSE} (Normalized Speaker Entropy): normalized entropy of speaker distribution. (0: high imbalance, 1: near uniform). $\uparrow$ score means $\uparrow$ balance. \\
& \textbf{SC-Gini} (Semantic Concentration): Gini coefficient of speaker contributions.
$\uparrow$ score means $\uparrow$ centralized; $\downarrow$ score means $\uparrow$ collaborative. \\

\midrule
\multirow{6}{*}{Content}
& \textbf{$\Phi$} (Task Success Indicator): binary indicator of whether the conversation satisfies pre-defined goals and success conditions. \\
& \textbf{ACR} (Agenda Completion Rate): fraction of objectives or agenda items that are fully addressed and covered by the conversation. \\
& \textbf{PE} (Progression Efficiency): average number of conversation turns needed to fully address and cover one agenda item. $\uparrow$ score means $\uparrow$ turns.\\
& \textbf{CS} (Conversational Structure): measures how consistently the conversation's agenda progression follows a reference order. \\
& \textbf{PD} (Progression Distance): semantic distance between the first and last utterances, capturing the overall topical movement in the conversation. \\
& \textbf{HMP} (Harmonic Mean Progression): harmonic mean of semantic distances between turns. $\uparrow$ score means $\uparrow$ steady flow. \\
\midrule
\multirow{1}{*}{\makecell{Speaker--\\Content}}
& \textbf{GSCC-DC} (Centroid-based GSCC): embedding similarity between a speaker's utterances and their semantic centroids (semantic clusters of the speaker's utterances). $\uparrow$ score means $\uparrow$ tightly clustered around central themes, reflecting higher role and topical consistency. \\
\bottomrule
\end{tabularx}
}
\vspace{3em}
\end{table*}
\section{Local Evaluation}
\label{sec:local-eval}

This section presents the design and rationale of \emph{local evaluation} in \textsc{MPCEval}. Local evaluation assesses the quality of a single predicted next turn and is designed to be both \emph{reference-free}---requiring no human-authored ground truth---and \emph{reproducible}---yielding deterministic scores given the same context and evaluation object. A summary of all local metrics is provided in Table~\ref{tab:mpceval_metrics}, with full technical definitions and implementation details deferred to Appendix~\ref{sec:local-old}.

We represent a conversation as a sequence of turns, where each turn is a speaker-utterance pair. Let the full conversation history be
$C=\langle (s_1,u_1),\ldots,(s_m,u_m)\rangle$, and let
$C_{-k}=\langle (s_{m-k+1},u_{m-k+1}),\ldots,(s_m,u_m)\rangle$
denote the most recent $k$ turns. Given $C_{-k}$, a model predicts the next turn $(s_{m+1},u_{m+1})$, including a speaker $s_{m+1}$ and message content $u_{m+1}$. \textbf{Local evaluation} assesses whether this predicted turn is appropriate given the immediate conversational context.

Local evaluation in \textsc{MPCEval} is decomposed into three complementary dimensions: \emph{next-speaker modeling}, \emph{local content quality}, and \emph{speaker--content consistency}. This decomposition reflects three failure modes in multi-party dialogue: selecting an implausible speaker, generating incoherent or redundant content, and attributing contents to inappropriate speakers. Each dimension is quantified by a small set of complementary metrics, designed to provide diagnostic insights rather than a single aggregated score.

\subsection{Local Speaker Modeling}

Local speaker modeling evaluates whether $s_{m+1}$ is a plausible next participant given recent conversational cues. In multi-party settings, speaker selection depends on multiple factors, including explicit addressing, participation recency, and topical relevance; no single cue is sufficient in isolation. \textsc{MPCEval} therefore models speaker plausibility through a small set of complementary signals.

When a speaker is explicitly addressed in the recent context, they are typically the most likely next speaker. \emph{Direct Name Reference} (DNR) captures this cue as a binary indicator that assigns a score of 1 if $s_{m+1}$ is explicitly mentioned (e.g., via ``@'' mentions) in $C_{-k}$, and 0 otherwise. When explicit addressing is absent, speaker choice may instead be implied by conversational dynamics or topic relevance.

To capture such cases, we introduce metrics based on (1) \emph{implicit turn-taking cues} and (2) \emph{topic-role fit}. \emph{Implicit Reference} (IR) measures whether $s_{m+1}$ has participated recently, reflecting engagement-based turn-taking patterns. \emph{Local Speaker Topic Alignment} (LS-TA) measures whether $s_{m+1}$ is topically aligned with the current discussion even if they have not spoken recently. Together with participation frequency and embedding-based similarity, these metrics form a diagnostic profile that distinguishes recency-driven from topic-driven speaker selection.

By separating these cues, \textsc{MPCEval} reveals which conversational signals a model relies on for speaker prediction rather than conflating them into a single-dimensional score such as prediction accuracy. All local speaker modeling metrics are summarized in Table~\ref{tab:mpceval_metrics}, with full definitions in Appendix~\ref{sec:local_speaker_measures}.

\subsection{Local Content Quality}

Local content quality evaluates whether $u_{m+1}$ is a relevant and informative continuation of the recent discussion. A desirable response should neither repeat prior content verbatim nor introduce content disconnected from the current conversational state.

We begin with \emph{Lexical Novelty Rate} (LNR), which measures the proportion of lexical units in $u_{m+1}$ that do not appear in $C_{-k}$. Extremely low values indicate redundancy, while extremely high values suggest topic drift. However, lexical novelty alone is insufficient, as a response may paraphrase existing content without adding information or introduce unrelated content using familiar words.

To address these limitations, \textsc{MPCEval} includes complementary measures that assess novelty and coherence beyond surface form. \emph{Embedding-aware Lexical Novelty} (LNR-E) and \emph{Inverse Document Frequency-Weighted LNR-E} (LNR-E-w) evaluate semantic novelty while reducing sensitivity to paraphrasing and synonyms. \emph{Dialogue-Act Transition Fit} (DAF) assesses whether $u_{m+1}$ is functionally appropriate given recent dialogue acts (e.g., answering a previously raised question). \emph{Topic Expansion Score} (TES) measures how smoothly $u_{m+1}$ maintains or extends the current topic, while \emph{Agenda Progression} (AP) evaluates advancement towards the end goal when explicit conversation objectives are available.

These metrics are intended to be interpreted jointly. For example, high LNR but low TES indicates lexical novelty with less topical changes, while low LNR with high DAF suggests functional appropriateness with potential redundancy. This decomposition exposes failure modes of local content continuation that would be obscured by a single fluency or coherence score. All local content quality metrics are listed in Table~\ref{tab:mpceval_metrics}, with details in Appendix~\ref{sec:local_content_measures}.

\subsection{Local Speaker--Content Consistency}

Local speaker--content consistency evaluates whether $u_{m+1}$ is consistent with $s_{m+1}$ under the local context. In multi-party conversations, participants often exhibit stable topical preferences, roles, or speaking styles; thus, even a plausible speaker choice and coherent message can form an implausible pair if misaligned.

To capture this phenomenon, we adopt \emph{Local Speaker--Content Consistency via Embedding Similarity} (LSCC-ES), which measures the semantic alignment between $u_{m+1}$ and $s_{m+1}$'s recent utterances in $C_{-k}$. We report both \emph{LSCC-ES-avg}, capturing overall consistency, and \emph{LSCC-ES-max}, capturing best-match consistency. When speaker profiles or background metadata are available, augmented variants improve robustness for speakers with sparse histories.

The variants provide complementary signals: high LSCC-ES-max but low LSCC-ES-avg indicates alignment with a single utterance but inconsistency with broader behavior. All local speaker--content consistency metrics are summarized in Table~\ref{tab:mpceval_metrics} with implementation and technical details listed in Appendix~\ref{sec:local_speaker_content_consistency}.

\emph{Synthesis.} Taken together, local evaluation in \textsc{MPCEval} provides a fine-grained assessment of next-turn quality by jointly analyzing speaker plausibility, content appropriateness, and speaker--content alignment. Reporting these dimensions separately enables diagnostic analysis, exposes systematic trade-offs, and avoids the misleading simplification of single-score evaluation, while remaining fully reference-free and reproducible.

\section{Global Evaluation}
\label{sec:global-eval}

This section presents the design and rationale of \emph{global evaluation} in \textsc{MPCEval}. Global evaluation assesses the quality of an entire generated conversation rather than individual turns and is designed to be \emph{reference-free}, \emph{quantitative}, and \emph{reproducible}. A summary of all global evaluation metrics is provided in Table~\ref{tab:mpceval_metrics}, with full definitions and implementation details deferred to Appendix~\ref{sec:global-old}.

Global evaluation targets long-range properties of multi-party interaction that cannot be captured at the turn level, including participation structure, information distribution and progression, and speaker coherence over extended context. We organize global evaluation into three complementary dimensions---global speaker modeling, global content quality, and global speaker--content consistency---each measured by a small set of metrics that jointly form a diagnostic view of the generated conversation's quality.
\vspace{-3mm}

\subsection{Global Speaker Modeling}

Global speaker modeling evaluates how speakers participate over the full interaction and how their contributions shape the conversation. Unlike local speaker modeling, which focuses on immediate next-speaker plausibility, global measures characterize long-range participation patterns and information concentration.

To quantify participation balance, we compute \emph{Normalized Speaker Entropy} (NSE), which measures how evenly turns are distributed across participants. Lower NSE indicates dominance by a few speakers, while higher NSE reflects more balanced participation. Balanced participation is not assumed to be universally optimal; rather, NSE serves as a descriptive signal that must be interpreted in context.

Moreover, balanced turn-taking does not imply balanced information contribution. To capture this distinction, we additionally report semantic concentration measures based on each speaker's contribution to the conversation's overall semantic span. In particular, the \emph{Semantic Concentration Gini} (SC-Gini) score measures how unevenly substantive information is distributed across speakers. Together, NSE and SC-Gini form a complementary diagnostic profile---for example, high NSE and high SC-Gini indicate balanced turn-taking with information concentrated in a few speakers.

These global speaker measures reveal participation structures that emerge over long horizons, rather than simple aggregations of local speaker choices. All global speaker modeling metrics are summarized in Table~\ref{tab:mpceval_metrics}, with detailed definitions in Appendix~\ref{sec:global_speaker_measures}.

\subsection{Global Content Quality}

Global content quality evaluates whether the generated conversation makes meaningful progress over time. Unlike local content quality, which assesses immediate coherence, global content quality captures long-range semantic development and progression.

When an explicit objective or agenda is available, progress is defined relative to that objective. We evaluate both \emph{whether} the conversation achieves the intended outcome and \emph{how} it does so. The \emph{Task Success Indicator} ($\Phi$) measures whether predefined success conditions are met, while \emph{Progression Efficiency} (PE) evaluates how efficiently the conversation advances toward completing the agenda, penalizing unnecessarily prolonged or circular discussion. These metrics are applied only when objectives or agenda structures are available or reliably inferred.

When no explicit objective is provided, we characterize progress through the conversation's semantic trajectory. \emph{Progression Distance} (PD) measures the net semantic displacement from the beginning to the end of the conversation, normalized by length. 
Low PD may indicate stagnation, while extremely high PD may signal arbitrary topic switching. 
To complement PD, we report \emph{Harmonic Mean Progression} (HMP), which favors smooth and steady semantic progression. Together, PD and HMP distinguish sustained development from uncontrolled topic shifts.

The global content metrics capture interaction-level progression patterns that emerge only over extended context. All global content quality metrics are listed in Table~\ref{tab:mpceval_metrics}, with details in Appendix~\ref{sec:global_content_measures}.

\subsection{Global Speaker--Content Consistency}

Global speaker--content consistency evaluates whether speakers behave consistently with established roles, expertise, or topical focus throughout the conversation. Even when local speaker choices and content appear reasonable, a conversation may be globally implausible if speakers' behavior shifts unpredictably over time.

To capture long-term consistency, we model each speaker's contributions as one or more centroids (i.e., semantic clusters), each representing a behavioral characteristics such as topical focus or functional role. \emph{Global Speaker--Content Consistency via Distance to Centroid} (GSCC-DC) is then a natural measure of how well a speaker's utterances align with these cluster representations over the full interaction. We report both \emph{GSCC-DC-avg}, capturing overall alignment, and \emph{GSCC-DC-max}, capturing best-match alignment.

When speaker profiles or background metadata are available, we incorporate this additional information into the cluster representations, allowing consistency to reflect alignment with both observed interactions and known background. When such metadata are unavailable or sparse, \textsc{MPCEval} derives speaker-specific clusters directly from the conversation, enabling robust consistency measurement without external supervision.

By modeling speaker behavior over long horizons, these metrics capture patterns that cannot be observed at the turn level or through simple aggregation of local scores. All global speaker--content consistency metrics are in Table~\ref{tab:mpceval_metrics} and details in Appendix~\ref{sec:global_speaker_content_consistency}.

\emph{Synthesis.} Taken together, global evaluation in \textsc{MPCEval} complements local evaluation by revealing long-range interaction properties invisible at the turn level. The proposed metrics jointly characterize participation structure, information distribution, semantic progression, and long-term speaker behavioral consistency, enabling principled assessment of multi-party conversation generation beyond aggregated local scores. As with local evaluation, all global metrics are reference-free and reproducible.

\section{Empirical Studies}\label{sec:experiments}

Our experiments are designed to systematically evaluate both the behavior of modern multi-party conversation generation models and the effectiveness of \textsc{MPCEval} as an evaluation framework. Specifically, we organize our empirical study around the following research questions, each targeting a distinct aspect of evaluation validity and interpretability.

\textbf{RQ1.} Do different generation models exhibit systematically different behaviors across speaker modeling, content quality, and speaker--content consistency, and can \textsc{MPCEval} reliably distinguish these behaviors across datasets and evaluation granularities (local vs.\ global)?

\textbf{RQ2.} How do the metrics introduced in \textsc{MPCEval} differ from widely used existing evaluation metrics in terms of sensitivity, diagnostic power, and robustness, and in what concrete scenarios do existing metrics fail to capture important dimensions of multi-party conversational quality that \textsc{MPCEval} explicitly measures?

\textbf{RQ3.} How do human-authored and machine-generated multi-party conversations differ across multiple evaluation dimensions, and to what extent should human-authored conversations be treated as a universal gold standard versus one point in a broader space of conversational trade-offs captured by \textsc{MPCEval}?

\subsection{Experiment Setup}
A central design principle of \textsc{MPCEval} is task awareness. Multi-party conversation generation encompasses qualitatively different generation objectives, and evaluation criteria must be aligned accordingly. Thus, as highlighted in Section~\ref{sec:measure_objective}, we explicitly distinguish two evaluation tasks---\emph{next-message prediction} and \emph{full-conversation generation}---and benchmark representative datasets and state-of-the-art (SOTA) generation methods under each setting.

\subsubsection{Datasets}
We evaluate \textsc{MPCEval} on three complementary multi-party conversation datasets introduced in Section~\ref{sec:dataset-method}, each having distinct characteristics to cover diverse real-life scenarios:
(1) \textbf{DeliData}~\cite{karadzhov2023delidata}, a collaborative problem-solving dataset, which supports both next-message prediction and full-conversation generation from scratch given topic constraints;
(2) \textbf{MPDD}~\cite{chen2020mpdd}, a public multi-party dialogue dataset derived from TV scripts, which is primarily suited for next-message prediction due to its shorter context length; and
(3) \textbf{Tanka}, a real-world enterprise communication dataset featuring long, information-dense interactions, which is particularly suitable for evaluating full-conversation generation (continual generation given conversation history).

Using multiple datasets allows us to assess whether observed findings, such as model behaviors and metric patterns, are consistent across public and private data, short and long conversations, and objective-driven versus open-ended interaction settings.

\subsubsection{Generation Methods}\quad

\noindent\textbf{Next-Message Prediction (Local).}
For local evaluation, we benchmark three categories of representative approaches:
(1) \textbf{MultiLIGHT}~\cite{wei2023multi}, a supervised training-based method;
(2) \textbf{ChatGPT-solver}~\cite{tan2023chatgpt}, a prompting-based approach tailored to \textsc{GPT-4}; and
(3) \textbf{Model-Agnostic Prompting}, where we evaluate a range of SOTA LLMs---including \textsc{Llama-3.3-70B}~\cite{dubey2024llama}, \textsc{GPT-4-Turbo}~\cite{achiam2023gpt}, \textsc{DeepSeek-V3}~\cite{liu2024deepseek}, and \textsc{Claude-3.5-Sonnet}~\cite{Anthropic2024Claude35}---using a unified prompting template.
This setup enables controlled comparison across models with different architectures and training paradigms while holding the prompting interface fixed. Detailed prompts and method descriptions are provided in Appendix~\ref{sec:appendix_detailed_generation_methods} and Appendix~\ref{app:prompt_prediction}.

\textbf{Full-Conversation Generation (Global).}
For global evaluation, we focus on model-agnostic prompting frameworks capable of generating multi-turn, multi-speaker interactions:
(1) \textbf{\textsc{MPC-constraints}}~\cite{penzo2025don}, a constraint-driven prompting method originally validated on \textsc{Llama-3.1}~\cite{dubey2024llama} and \textsc{Qwen-2.5}~\cite{qwen25}; and
(2) an expansion of \textsc{MPC-constraints} to additional SOTA models, including \textsc{GPT-4-Turbo}, \textsc{Llama-3.3-70B}, \textsc{DeepSeek-V3}, and \textsc{Claude-3.5-Sonnet}.
Detailed method descriptions are in Appendix~\ref{sec:appendix_detailed_generation_methods}.

\textbf{Scope and Limitations.}
Several additional multi-party conversation generation methods have been proposed in prior work~\cite{li2024chatmdg,chernyavskiy2023transformer,gu2022hetermpc,mahajan2024persona}. However, these methods typically rely on auxiliary information such as explicit addressing annotations, speaker profiles, or discourse structures that are unavailable or inconsistently annotated in most large-scale datasets. To ensure fair comparison and broad applicability, we restrict our experiments to methods that operate using only raw conversation history and optional task descriptions and speaker information. Importantly, this restriction reflects current data availability rather than a limitation of \textsc{MPCEval} itself: as validated by the more sophisticated metrics in \textsc{MPCEval}, which incorporate profile information when available, while still providing effective, simpler variants when such additional information is unavailable, the framework is explicitly designed to accommodate additional metrics and evaluation dimensions---such as addressing accuracy, persona consistency, or affective alignment---when richer annotations become available.

\begin{table*}[t]
\centering
\caption{Performance comparison on the DeliData dataset using \textsc{MPCEval}'s next-message prediction (local) measures (mean $\pm$ standard deviation). \textbf{Bold} and \underline{underlined} values indicate the highest and lowest scores in each column, respectively. For \textsc{LNR-E-w}, M-SNS-avg, and \textsc{TES}, values closest to 0.5 are \textbf{bolded}, while those farthest from 0.5 are \underline{underlined}. \textsc{ChatGPT-solver} does not report LSCC-ES-avg, as local speaker--content consistency is not applicable (see Appendix~\ref{subsec:appendix_local}).\vspace{-3mm}}
\label{tab:delidata_local_all}
\small
\setlength{\tabcolsep}{2.5pt}
\resizebox{0.85\textwidth}{!}{%
\begin{tabular}{lccccc|ccccc|c}
\toprule
& \multicolumn{5}{c|}{\textbf{Local Speaker Modeling}} 
& \multicolumn{5}{c|}{\textbf{Local Content Quality}} 
& \multicolumn{1}{c}{\textbf{Local Consistency}} \\
\cmidrule(lr){2-6}\cmidrule(lr){7-11}\cmidrule(lr){12-12}
\textbf{Model} 
& \textbf{DNR} & \textbf{IR} & \textbf{PF} & \textbf{LS-ES-avg} & \textbf{LS-TA}
& \textbf{LNR-E-w} & \textbf{M-SNS-avg} & \textbf{DAF} & \textbf{LL} & \textbf{TES}
& \textbf{LSCC-ES-avg} \\
\midrule
\textsc{LLaMa-3.3}
& .278$\pm$.448 & .182$\pm$.205 & \underline{.276$\pm$.096} & \underline{.444$\pm$.088 }& .778$\pm$.193
& .280$\pm$.159 & .735$\pm$.091 & .281$\pm$.243 & .893$\pm$.091 & .357$\pm$.399
& .358$\pm$.198 \\
\textsc{GPT-4-Turbo}
& .278$\pm$.448 & .185$\pm$.203 & .303$\pm$.100 & .456$\pm$.097 & .783$\pm$.193
& \textbf{.403$\pm$.177} & .736$\pm$.077 & \textbf{.363$\pm$.237} & .793$\pm$.178 & .142$\pm$.168
& \textbf{.421$\pm$.206} \\
\textsc{DeepSeek}
& \textbf{.389$\pm$.488} & .213$\pm$.223 & .278$\pm$.099 & .450$\pm$.100 & \underline{.774$\pm$.196}
& .314$\pm$.133 & \textbf{.688$\pm$.061} & .307$\pm$.240 & \textbf{.918$\pm$.074} & \underline{.112$\pm$.185}
& .381$\pm$.161 \\
\textsc{Claude-3.5}
& .333$\pm$.471 & \underline{.125$\pm$.143} & .298$\pm$.101 & .454$\pm$.096 & .836$\pm$.202
& .343$\pm$.176 & .698$\pm$.081 & \underline{.279$\pm$.236} & .838$\pm$.203 & \textbf{.500$\pm$.497}
& .331$\pm$.178 \\
\midrule
\textsc{MultiLIGHT}
& .333$\pm$.471 & .227$\pm$.216 & .309$\pm$.102 & .448$\pm$.088 & .781$\pm$.174
& \underline{.156$\pm$.233} & \underline{.749$\pm$.072} & .329$\pm$.267 & .757$\pm$.222 & .303$\pm$.407
& \underline{.297$\pm$.211} \\
\textsc{ChatGPT-solver}
& .333$\pm$.471 & .195$\pm$.196 & \textbf{.369$\pm$.126} & \textbf{.489$\pm$.098} & \textbf{.848$\pm$.160}
& .289$\pm$.202 & .690$\pm$.068 & .294$\pm$.238 & .833$\pm$.192 & .167$\pm$.373
& -- \\
\midrule
Human
& \underline{.222$\pm$.416 }& \textbf{.295$\pm$.254} & .312$\pm$.110 & .458$\pm$.103 & .816$\pm$.191
& .245$\pm$.299 & .719$\pm$.109 & .348$\pm$.234 & \underline{.232$\pm$.349} & .281$\pm$.383
& .334$\pm$.203 \\
\bottomrule
\end{tabular}
}
\end{table*}

\begin{figure*}[t]
\centering
\small

\setlength{\tabcolsep}{4pt}

\fcolorbox{black}{gray!10}{%
  \parbox{\dimexpr\textwidth-2\fboxsep-2\fboxrule\relax}{%
    \scriptsize
    \textbf{Context:} This conversation is from the DeliData dataset, where participants collaboratively solve the Wason card selection task. In this specific instance, four cards are presented showing: \textbf{5}, \textbf{6}, \textbf{N}, and \textbf{A}. Each card has a letter on one side and a number on the other. Participants must determine which cards to flip to test the rule: ``If a card has a vowel on one side, it has an even number on the other side.'' The participants are debating which cards to select to verify or falsify this rule.
  }%
}
\vspace{4pt}

\begin{tabularx}{\textwidth}{|>{\raggedright\arraybackslash}X|>{\centering\arraybackslash}p{0.38\textwidth}|}
\hline
\multicolumn{1}{|c|}{\textbf{Conversation History \& Next Message}} &
\multicolumn{1}{c|}{\textbf{Metric Comparison}} \\
\hline

\scriptsize
\textbf{Tiger:} So what do you guys think \newline
\textbf{Ox:} I would turn A and 6, but it can't be that easy \newline
\textbf{Falcon:} I was thinking the same too, but we need to discuss \newline
\textbf{Falcon:} I think turning A would be a no-brainer \newline
\textbf{Falcon:} @Tiger, what is your opinion \newline
\textbf{Tiger:} I was thinking, maybe we should turn all odd and even numbers
\vspace{4pt}

\textbf{Generated Next Message:}\newline
\textbf{Ox:} Actually, we only need to turn the A and 5. A to check if there's an even number on the other side, and 5 to ensure there's no vowel on its other side.
\vspace{2pt}

\textbf{Human-Authored Next Message:}\newline
\textbf{Ox:} hm.. why the odd numbers
&
\scriptsize
\vspace{1pt}

\renewcommand{\arraystretch}{1.25}
\begin{tabular}{@{}l>{\centering\arraybackslash}c>{\centering\arraybackslash}c|l>{\centering\arraybackslash}c>{\centering\arraybackslash}c@{}}
\toprule
\textbf{Existing Metric} & \textbf{Gen.} & \textbf{Human} &
\textbf{\textsc{MPCEval}} & \textbf{Gen.} & \textbf{Human} \\
\midrule
BLEU-4         & 0.006  & ---    & LNR-E-w   & 0.591 & \textbf{0.469} \\
BERTScore (F1) & 0.840  & ---    & M-SNS-avg & 0.774 & \textbf{0.673} \\
BARTScore      & \textbf{-3.073} & -5.983 & M-SNS-min & \textbf{0.490} & 0.327 \\
G-EVAL (Engage.)  & \textbf{2.700}  & 1.000  & DAF       & \textbf{0.700} & 0.000 \\
Uni-Eval (Engage.)               &   \textbf{1.995}     &   0.985     & LL        & \textbf{0.956} & 0.906 \\
Uni-Eval (Understand.)            &   0.290     &   \textbf{0.530}     & TES       & \textbf{0.516} & 0.000 \\
\bottomrule
\end{tabular}
\\
\hline
\end{tabularx}\vspace{-3mm}
\caption{Case study comparing \textsc{MPCEval} with existing evaluation metrics. ``Gen.'' denotes model-generated messages. BLEU-4 and BERTScore are excluded for human-authored next message as comparing the reference to itself is uninformative.}
\label{fig:case-study-RQ2}
\end{figure*}

\subsection{RQ1: Comparison of Generation Models}

Tables~\ref{tab:delidata_local_all} and~\ref{tab:global_delidata_final} show that different generation models exhibit systematically different behaviors across speaker modeling, content quality, and speaker--content consistency dimensions, and that \textsc{MPCEval} effectively differentiates these behaviors at both local and global levels. For brevity, we present detailed results on the DeliData dataset in the main paper and report complete results across all datasets in Appendix~\ref{sec:additional_experiment}, where consistent trends are observed.

\begin{table}[t]
\centering
\caption{Performance on the DeliData dataset via \textsc{MPCEval}'s global metrics (mean $\pm$ std).
GSCC denotes GSCC-DC-avg under single-centroid speaker--content consistency. \textbf{Bold} and \underline{underlined} mark the highest and lowest values per column. \textsc{MPC} is shorthand for the \textsc{MPC-constraints} method~\cite{penzo2025don}.\vspace{-2mm}}
\label{tab:global_delidata_final}
\small
\setlength{\tabcolsep}{2.8pt}
\resizebox{\columnwidth}{!}{
\begin{tabular}{l cc cc c}
\toprule
& \multicolumn{2}{c}{\textbf{Global Speaker}} & \multicolumn{2}{c}{\textbf{Global Content}} & \textbf{Consistency} \\
\cmidrule(lr){2-3} \cmidrule(lr){4-5} \cmidrule(lr){6-6}
\textbf{Model} & \textbf{NSE} & \textbf{SC-Gini} & \textbf{PD} & \textbf{HMP} & \textbf{GSCC} \\
\midrule
\textsc{MPC + LLaMa-3.1}  & .978$\pm$.022 & \textbf{.445$\pm$.197} & 1.127$\pm$.163 & .869$\pm$.107 & .894$\pm$.040 \\
\textsc{MPC + Qwen}       & .983$\pm$.024 & .392$\pm$.176 & \textbf{1.648$\pm$.215} & \textbf{1.021$\pm$.114} & \textbf{.951$\pm$.061} \\
\midrule
\textsc{MPC + LLaMa-3.3}  & \underline{.960$\pm$.000} & .430$\pm$.163 & 1.144$\pm$.148 & .835$\pm$.074 & .915$\pm$.026 \\
\textsc{MPC + GPT-4-Turbo}    & .984$\pm$.020 & .420$\pm$.185 & 1.198$\pm$.232 & .875$\pm$.095 & .885$\pm$.039 \\
\textsc{MPC + DeepSeek}   & .979$\pm$.024 & .416$\pm$.197 & 1.270$\pm$.145 & \underline{.807$\pm$.095} & .918$\pm$.041 \\
\textsc{MPC + Claude-3.5} & \textbf{.987$\pm$.015} & .366$\pm$.203 & 1.280$\pm$.115 & .967$\pm$.070 & .862$\pm$.036 \\
\midrule
Human                     & .962$\pm$.024 & \underline{.245$\pm$.182} & \underline{.850$\pm$.198} & .873$\pm$.484 & \underline{.528$\pm$.065} \\
\bottomrule
\end{tabular}
}
\end{table}

\textbf{Local speaker modeling.}
Local speaker modeling results (Table~\ref{tab:delidata_local_all}) indicate that models rely on distinct conversational cues for next-speaker prediction. \textsc{DeepSeek} excels in Direct Name Reference (DNR\,=\,0.389), reflecting a stronger capability in detecting explicit addressee mentions. In contrast, \textsc{ChatGPT-solver} performs best on Participation Frequency (PF\,=\,0.369), Local Speaker Embedding Similarity (LS-ES-avg\,=\,0.489), and Local Speaker Topic Alignment (LS-TA\,=\,0.848), suggesting greater emphasis on participation patterns and topical continuity. These differences illustrate how \textsc{MPCEval} differentiates multiple speaker-selection strategies that would otherwise be conflated by single aggregate metrics.

\textbf{Global content quality.}
Global content quality metrics (Table~\ref{tab:global_delidata_final}) reveal models' distinct long-range progression styles in content generation. \textsc{Claude-3.5-Sonnet} attains the highest Progression Distance (PD\,=\,1.280) and Harmonic Mean Progression (HMP\,=\,0.967), indicating both substantial semantic advancement and smooth turn-to-turn progression. \textsc{DeepSeek}, while achieving a comparable PD (1.270), exhibits a lower HMP (0.807), reflecting more uneven progression with intermediate stagnation or abrupt semantic shifts. These results demonstrate that similar aggregate progress can arise from qualitatively different interaction dynamics, which are explicitly captured by \textsc{MPCEval}'s decomposed global metrics.

Overall, \textsc{MPCEval} reveals clear, dimension-specific model behaviors across both local and global settings, enabling more nuanced and diagnostic comparison of multi-party conversation generation systems than single-score evaluations.

\subsection{RQ2: Evaluation Metrics in \textsc{MPCEval} vs.\ Existing Representative Metrics}
We conduct a case study on the DeliData dataset to compare \textsc{MPCEval}'s metrics with representative metrics, including BLEU~\cite{papineni2002bleu}, BERTScore~\cite{zhang2019bertscore}, BARTScore~\cite{yuan2021bartscore}, G-Eval~\cite{liu2023g}, and Uni-Eval~\cite{zhong2022towards}.

As indicated in Figure~\ref{fig:case-study-RQ2}, reference-based metrics such as BLEU and BERTScore assign low scores to the model-generated response because it diverges from the single human-authored reference. However, the generated turn remains coherent and contextually appropriate, illustrating a fundamental limitation of reference-based evaluation in the open-ended setting of multi-party dialogue, where multiple continuations can be equally valid.

Existing reference-free metrics also exhibit notable limitations. BARTScore assigns a very low likelihood to the human-authored turn, even though low likelihood does not necessarily imply incoherence in conversational contexts. Similarly, LLM-as-a-judge methods such as G-Eval and Uni-Eval assign high engagement scores to the generated response while penalizing the human-authored one, despite both introducing comparably novel and relevant ideas. Uni-Eval also assigns a low understandability score to the generated turn, contradicting qualitative inspection. These discrepancies highlight the limited diagnostic power of existing metrics.

In contrast, \textsc{MPCEval} provides a set of quantitative, reference-free, and reproducible metrics that explicitly decompose conversational quality into complementary dimensions. For example, lower DAF and LL scores for human-authored turns reflect reduced conformity to common dialogue-act patterns or language-model expectations, while LNR-E-w and M-SNS-avg still capture their higher semantic novelty and contextual relevance. Together, these metrics offer a more balanced and interpretable assessment of conversational quality than single-score alternatives.

\subsection{RQ3: Human vs.\ Generated Conversation}

Tables~\ref{tab:delidata_local_all} and~\ref{tab:global_delidata_final} show that human-authored and machine-generated conversations exhibit systematically different conversational characteristics rather than forming a strict quality hierarchy.

At the local level (Table~\ref{tab:delidata_local_all}), human-authored conversations achieve the highest Implicit Reference (IR\,=\,0.295 vs.\ a machine-generated average of 0.188), but the lowest Direct Name Reference (DNR\,=\,0.222 vs.\ 0.324) and Log-Likelihood (LL\,=\,0.232 vs.\ 0.839). This pattern indicates greater reliance on implicit turn-taking cues and lower predictability in human utterances. Human-authored conversations also show lower \textsc{TES} (0.281) and \textsc{LNR-E-w} (0.245), reflecting conservative topical progression and higher lexical redundancy. In contrast, machine-generated dialogues---particularly those from \textsc{GPT-4-Turbo} and \textsc{Claude-3.5-Sonnet}---closely approach the novelty-coherence balance, with \textsc{TES} values near 0.500.

At the global level (Table~\ref{tab:global_delidata_final}), both human and machine-generated conversations exhibit high participation diversity (NSE), but human-authored dialogues show substantially lower global speaker--content consistency (GSCC\,$\approx$\,0.53) compared to generated ones (GSCC\,$\approx$\,0.86--0.95). Machine-generated conversations also demonstrate stronger and steadier semantic progression, as indicated by higher PD and HMP values, whereas human-authored conversations evolve more conservatively.

Overall, these results suggest that human-authored conversations should not be treated as a gold standard. Instead, human and machine-generated dialogues emphasize complementary dimensions of conversational quality, which \textsc{MPCEval} makes explicit.

\section{Conclusion and Future Works}\label{sec:conclusion}

We introduced \textsc{MPCEval}, a task-aware and decomposed evaluation framework for multi-party conversation generation. By separating speaker modeling, content quality, and speaker--content consistency, and by distinguishing local next-message prediction from global conversation generation, \textsc{MPCEval} provides quantitative, reference-free, and reproducible metrics that capture interaction properties missed by existing evaluations. Experiments show clear, dimension-specific behaviors across models, highlight limitations of widely used metrics in multi-party settings, and demonstrate that human-authored conversations are not a universal gold standard.

Future work will extend \textsc{MPCEval} to incorporate richer annotations such as explicit addressing or affective signals, analyzing metric robustness under different modeling choices, and integrating controlled human evaluations to better calibrate automatic metrics. We also plan to study the scalability and computational cost of the framework on longer, larger-scale conversations and to release additional public benchmarks to facilitate broader adoption.


\bibliographystyle{ACM-Reference-Format}
\bibliography{sample-base}

\newpage
\appendix
\section{Dataset Details}\label{sec:dataset-detail}
We detail the characteristics of each dataset and their intended evaluation tasks below.

\textbf{Delidata}~\cite{karadzhov2023delidata}.  
A publicly-available collaborative problem-solving dataset based on the Wason selection task, containing 500 dialogues (avg.\ 27--40 utterances, avg.\ 3--4 speakers). Because each dialogue has an explicit objective and structured setup (Wason selection task initial setup), Delidata supports both \emph{next-message prediction} task (local evaluation) and \emph{full-conversation generation} task (global evaluation) by providing only the task description and speaker list as input. Under the full-conversation generation setting, content-quality measures are most applicable, as speaker profiles and conversation histories are not available.

\textbf{MPDD}~\cite{chen2020mpdd}.  
A publicly-available Chinese multi-party dialogue dataset derived from TV scripts, annotated with emotions and interpersonal relations. After filtering short dialogues, it contains 1,774 conversations (avg.\ 10.3 utterances, avg.\ 2.5 speakers). MPDD is well suited for \emph{next-message prediction} task (local evaluation), where the final turn can be masked and evaluated. It is less suitable for full-conversation generation due to constrained conversation lengths and the absence of explicit topics for each dialogue.

\textbf{Tanka}.  
Tanka is a private dataset of real enterprise communication and knowledge management collected from the Tanka platform between 2020 and November 2024. It contains multilingual (Chinese and English) content arising from authentic organizational collaboration and knowledge-sharing activities. From this corpus, we selected four highly active groups and segmented them into 20 dialogue instances using topic-initiating trigger messages identified by message IDs. These dialogues were manually curated by Tanka annotators to ensure high information density and minimal off-topic chatter. The resulting dataset features longer and richer interactions than public benchmarks, with an average dialogue length of 163.0 utterances (std.\ 95.85, range 53--338) and an average of 9.11 distinct speakers per dialogue (std.\ 2.64, max 14). This makes Tanka well suited for both \emph{next-message prediction} task (local evaluation) and \emph{full-conversation generation} task (global evaluation). For global evaluation, we apply length-based filtering to exclude dialogues that are too short (fewer than 50 utterances) or excessively long (more than 450 utterances), ensuring sufficient context while remaining within practical generation limits.

\section{Detailed Local Evaluation} 
\label{sec:local-old}

\begin{table}[t]
\centering
\footnotesize
\setlength{\tabcolsep}{5pt} 
\renewcommand{\arraystretch}{1.2} 

\definecolor{headergray}{gray}{0.85}
\definecolor{subheadergray}{gray}{0.96}

\caption{Core notations used in Appendix~\ref{sec:local-old} and~\ref{sec:global-old}.}
\label{tab:core-notation}

\begin{minipage}{0.88\columnwidth}
\centering
\begin{tabularx}{\linewidth}{|lX|}
\hline

\rowcolor{headergray} 
\textbf{Notation} & \textbf{Description} \\ 
\hline

\rowcolor{subheadergray} 
\multicolumn{2}{|l|}{\textbf{Conversation primitives}} \\
$C$ & Conversation history. \\
$\tilde{C}$ & Generated conversation. \\
$\hat{C}=C\oplus\tilde{C}$ & Full conversation (or $\hat{C}=\tilde{C}$ if generated from scratch). \\
$\oplus$ & Concatenation of turn sequences. \\
$(s_i,u_i)$ & The $i$-th turn with speaker $s_i$ and utterance $u_i$. \\
$C_{-k}$ & Most recent $k$ turns from $C$. \\
\hline

\rowcolor{subheadergray} 
\multicolumn{2}{|l|}{\textbf{Speakers and turn subsets}} \\
$\mathcal{S}$, $|\mathcal{S}|$ & Set of speakers and its size. \\
$C_{s}$ & Subset of turns by $s$ in $C_{-k}$. \\
$\bar{C}{s}$ & Subset of turns not by $s$ in $C{-k}$. \\
$\hat{C}_s$ & Subset of turns by $s$ in $\hat{C}$. \\
\hline

\rowcolor{subheadergray} 
\multicolumn{2}{|l|}{\textbf{Embeddings and similarity}} \\
$e(\cdot)$ & Embedding function for text segments. \\
$\mathrm{sim}(\cdot,\cdot)$ & Similarity in embedding space (e.g., cosine similarity). \\
\hline

\rowcolor{subheadergray} 
\multicolumn{2}{|l|}{\textbf{Topic modeling}} \\
$\theta(\cdot)\in\mathbb{R}^L$ & Topic distribution over $L$ topics for a text segment. \\
$L$ & Number of topics in the topic model. \\
\hline

\rowcolor{subheadergray} 
\multicolumn{2}{|l|}{\textbf{Agenda items}} \\
$\mathcal{A}$, $|\mathcal{A}|$ & Collection of agenda items and its cardinality. \\
$a$ & An agenda item. \\
\hline

\rowcolor{subheadergray} 
\multicolumn{2}{|l|}{\textbf{Background augmentation}} \\
$B_s$ & (Optional) background profile for speaker $s$. \\
$\mathrm{aug}_s$ & Background-augmented representation of speaker $s$. \\
\hline

\end{tabularx}
\end{minipage}
\end{table}

\nop{
Local evaluation considers a single predicted turn and assesses how closely it matches the immediate conversational context. Let $C=\langle (s_1,u_1),(s_2, u_2),\ldots,(s_m,u_m)\rangle$ denote the conversation history, $C_{-k} = \langle (s_{m-k+1},u_{m-k+1}),\ldots,(s_m,u_m)\rangle$ its latest $k$ turns, $s_p$ the predicted speaker, and $u_p$ the predicted content. At the local level, the three measurement aspects introduced in Section~\ref{sec:benchmark-objective} are illustrated as follows: 
\begin{enumerate}
    \item \textbf{Local Speaker Modeling}, detailed in Section~\ref{sec:local_speaker_measures}, quantifies how $s_p$ aligns with near-term cues (e.g., @mentions, participation recency, role expertise) present in $C_{-k}$.
    \item \textbf{Local Content Quality}, detailed in Section~\ref{sec:local_content_measures}, measures how $u_p$ is an appropriate continuation of the discussion given $C_{-k}$.
    \item \textbf{Local speaker--content Consistency}, detailed in Section~\ref{sec:local_speaker_content_consistency}, links $u_p$ back to $s_p$ to capture how the predicted content matches the attributed speaker's profile or historical interactions.
\end{enumerate}
}

Local evaluation assesses a single predicted next turn given the immediate conversational context. We represent a conversation as a sequence of turns, where each turn is a speaker-utterance pair. Let
$C=\langle (s_1,u_1),(s_2,u_2),\ldots,(s_m,u_m)\rangle$ denote the conversation history, and let
$C_{-k}=\langle (s_{m-k+1},u_{m-k+1}),\ldots,(s_m,u_m)\rangle$ denote the most recent $k$ turns.
Given $C_{-k}$, a model predicts the next turn $(s_{m+1},u_{m+1})$. Local evaluation assesses whether this predicted turn is proper given the immediate context.

At the local level, the three measurement aspects introduced in Section~\ref{sec:benchmark-objective} specialize as follows:
\begin{enumerate}
    \item \textbf{Local Speaker Modeling} (Section~\ref{sec:local_speaker_measures}) evaluates whether the predicted speaker $s_{m+1}$ is a plausible next participant given near-term conversational cues in $C_{-k}$ (e.g., explicit addressing, participation patterns, and role/topic alignment).
    \item \textbf{Local Content Quality} (Section~\ref{sec:local_content_measures}) evaluates whether the predicted utterance $u_{m+1}$ is an appropriate continuation of the recent discussion given $C_{-k}$.
    \item \textbf{Local speaker--content Consistency} (Section~\ref{sec:local_speaker_content_consistency}) evaluates whether $u_{m+1}$ is consistent with the attributed speaker $s_{m+1}$, based on the speaker's profile and/or their historical interactions in $C$.
\end{enumerate}


\subsection{Local Speaker Modeling}
\label{sec:local_speaker_measures}

\textbf{Motivation:} Local speaker modeling focus on predicting the next speaker identity, which is a fundamental challenge in multi-party conversation modeling. Appropriate next-speaker prediction reflects the model's understanding of conversational dynamics, participant roles, and discussion context. For instance, if a speaker explicitly addresses another participant (e.g., via an @mention), the model should prioritize that participant as the next speaker. Conversely, predicting a participant whose role or expertise is clearly misaligned with the ongoing discussion indicates a failure to capture conversational structure and participant relevance.

\textbf{Definition:} Local speaker modeling evaluates whether the predicted speaker $s_{m+1}$ is justified given the conversation history $C$ and (when available) participant background information. In particular, it characterizes the relevance of $s_{m+1}$ to the current discussion by leveraging: (1) the speaker's conversational history $C_{s_{m+1}}=\{(s_i,u_i)\in C \mid s_i=s_{m+1}\}$, (2) their background information $B_{s_{m+1}}$ (e.g., profile and role) whenever available, where $\{B_s\}_{s\in S}$ denotes background information for all speakers, and (3) the current discussion theme reflected in the most recent messages $C_{-k}$.

These measures capture whether the predicted speaker exhibits \textit{domain expertise} (e.g., relevant knowledge expressed in prior turns, an established role related to the topic, or background information indicating expertise) and/or \textit{topic engagement} (e.g., active participation in the relevant thread).

We propose five metrics to quantify local speaker modeling: direct name reference, implicit reference, participation frequency, local speaker embedding-based similarity, and local speaker topic-based alignment.

\subsubsection{Direct Name Reference (DNR}
\nop{
One intuitive quantification method is whether the predicted speaker is directly referred in the previous $k$ messages, i.e., the speaker's name appears in direct address form, denoted as $\text{Addressed}(C_{-k})$. For instance, this includes explicit ``@'' mentions (e.g., ``@username'').

Given that $s_p$ denote the predicted speaker, $C$ denote the entire conversation history, and $C_{-k}$ represent the previous $k$ messages with respect to the predicted message. The direct name reference score is:
\begin{equation}
    \text{DNR} (s_p, C_{-k}) = 
    \begin{cases}
        1 & \text{if } s_p \in \text{Addressed}(C_{-k}) \\
        0 & \text{otherwise}
    \end{cases}
\end{equation}
}

A direct and intuitive cue for next-speaker prediction is whether the predicted speaker is explicitly addressed in the most recent context. Let $\text{Addressed}(C_{-k})$ denote the set of speakers who are directly referred to in the previous $k$ messages (e.g., via explicit ``@'' mentions such as ``@username''). The \emph{direct name reference} score is defined as:
\begin{equation}
\label{eq:dnr}
    \text{DNR}(s_{m+1}, C_{-k}) = 
    \begin{cases}
        1 & \text{if } s_{m+1} \in \text{Addressed}(C_{-k}), \\
        0 & \text{otherwise.}
    \end{cases}
\end{equation}
A higher DNR score of 1 indicates that $s_{m+1}$ is explicitly solicited in the immediate context, whereas $\text{DNR}=0$ indicates that no explicit addressing cue supports $s_{m+1}$ within $C_{-k}$.

\subsubsection{Implicit Reference (IR)}

The predicted next speaker $s_{m+1}$ may also be supported by \emph{implicit} turn-taking patterns, even when no explicit addressing cue is present. In multi-party conversations, speakers who participated more recently in the current thread are often more likely to speak again, with this likelihood decaying as their last participation becomes more distant.

We operationalize this intuition with a distance-based decay function. Let $s_{-i}$ denote the speaker of the $i$-th most recent turn in $C_{-k}$ (so $s_{-1}=s_m$), and we exclude the immediately previous speaker at position $-1$. For a candidate occurrence at position $-i$ with $2 \le i \le k$, we define its conversational distance as $d=i-2$ (thus the most recent eligible position $-2$ has $d=0$). If a speaker appears multiple times in $C_{-k}$, we take the speaker's \emph{most recent} participation. The \emph{implicit reference} score is:
\begin{equation}
\label{eq:ir}
    \text{IR}(s_{m+1}, C_{-k}) =
    \begin{cases}
        \displaystyle \max_{i \in \mathcal{I}(s_{m+1})} f(i-2) & \text{if } \mathcal{I}(s_{m+1}) \neq \emptyset, \\
        0 & \text{otherwise,}
    \end{cases}
\end{equation}
\vspace{-0.2em}
\noindent where $\mathcal{I}(s)=\{i | 2 \le i \le k,\; s_{-i}=s_{m+1}\}$.

Here, $f(\cdot)$ is a monotonically decreasing function, assigning higher scores to smaller distances (more recent participation). Typical choices include geometric decay $f(d)=\lambda(1-\lambda)^d$, exponential decay $f(d)=e^{-\alpha d}$, or inverse distance $f(d)=\frac{1}{1+d}$.

A higher IR score indicates that $s_{m+1}$ appeared more recently within $C_{-k}$, providing implicit support for selecting $s_{m+1}$ as the next speaker; $\text{IR}=0$ indicates that $s_{m+1}$ does not appear in $C_{-k}$.

\begin{example}
Consider the following sequence of turns:
\begin{itemize}
    \item Alice: ``What's the status of the API integration?''
    \item Bob: ``We're at 80\% completion.''
    \item Charlie: ``The authentication module is done.''
    \item Alice: ``Great! When can we test?''
    \item \textbf{Next message:} Who speaks?
\end{itemize}
Using geometric decay $f(d)=\lambda(1-\lambda)^d$ with $\lambda=0.6$, Charlie appears at position $-2$ (distance $d=0$) and receives score $0.6$, whereas Bob appears at position $-3$ (distance $d=1$) and receives score $0.24$; the last speaker (Alice at $-1$) is excluded. Thus, IR favors Charlie as the most recently engaged speaker before the last turn.
\end{example}

\subsubsection{Participation Frequency (PF)}

The predicted next speaker $s_{m+1}$ is also often supported by their recent engagement in the ongoing discussion. We quantify this engagement by the \emph{participation frequency}, defined as the fraction of turns in $C_{-k}$ spoken by $s_{m+1}$:
\begin{equation}
\label{eq:pf}
    \text{PF}(s_{m+1}, C_{-k}) =
    \frac{\left|\{(s,u)\in C_{-k} | s = s_{m+1}\}\right|}{k}.
\end{equation}
A higher PF indicates $s_{m+1}$ being active in the recent context, whereas a lower PF indicates limited participation.

\subsubsection{Local Speaker Embedding-based Similarity (LS-ES)}\label{sec:local-speaker-embedding}

Beyond near-term turn-taking cues (e.g., explicit addressing, recency, and participation), an additional signal for next-speaker appropriateness is whether the predicted speaker $s_{m+1}$ is topically aligned with the current discussion. We quantify this by measuring embedding similarity between the ongoing conversation $C_{-k}$ and messages previously sent by $s_{m+1}$. We consider two variants: average embedding similarity and maximum embedding similarity. Let $C_{s_{m+1}}=\{(s_i,u_i)\in C_{-k} | s_i=s_{m+1}\}$ denote the set of recent turns authored by $s_{m+1}$, and $\bar{C}_{s_{m+1}}=\{(s_i,u_i)\in C_{-k} | s_i\neq s_{m+1}\}$ denote the set of recent turns not authored by $s_{m+1}$. We use $e(\cdot)$ as an embedding function and $\text{sim}(\cdot,\cdot)$ as a similarity metric (e.g., cosine similarity); for a turn $(s,u)$, we abuse notation and write $e((s,u))=e(u)$.

\paragraph{Average Local Speaker Embedding Similarity (LS-ES-avg)}
\begin{equation}
\label{eq:ls-es-avg}
    \text{LS-ES-avg}(s_{m+1}, C_{-k}) =
    \operatorname*{avg}_{c \in \bar{C}_{s_{m+1}}}
    \max_{c' \in C_{s_{m+1}}}
    \text{sim}\!\left(e(c), e(c')\right).
\end{equation}

\paragraph{Maximum Local Speaker Embedding Similarity (LS-ES-max)}
\begin{equation}
\label{eq:ls-es-max}
    \text{LS-ES-max}(s_{m+1}, C_{-k}) =
    \max_{c \in \bar{C}_{s_{m+1}}}
    \max_{c' \in C_{s_{m+1}}}
    \text{sim}\!\left(e(c), e(c')\right).
\end{equation}
A higher LS-ES score indicates stronger semantic alignment between the current discussion and $s_{m+1}$'s prior messages. In particular, LS-ES-avg emphasizes \emph{overall} alignment across the recent context, whereas LS-ES-max captures whether there exists \emph{at least one} strong topical connection to $s_{m+1}$'s history.

\paragraph{Incorporating Background Information for Sparse Conversation History}
The embedding-based measures above rely on $C_{s_{m+1}}$. When $|C_{s_{m+1}}|$ is small (or empty), the similarity measures can be unreliable or undefined. To address this, if available, we incorporate background information $B_{s_{m+1}}$ of $s_{m+1}$ (e.g., profile description, role, or bio) to supplement sparse conversation history.

Let $\mathbf{v}_{s_{m+1}} = \operatorname{avg}_{c \in C_{s_{m+1}}} e(c)$ capture the semantics of $s_{m+1}$'s recent utterances, and $\mathbf{b}_{s_{m+1}} = e(B_{s_{m+1}})$ capture the semantics of $s_{m+1}$'s background information. We define an augmented speaker representation $\operatorname{aug}_{s_{m+1}}$ as a combination of $\mathbf{{v}_{s_{m+1}}}$ and $\mathbf{b}_{s_{m+1}}$:
\begin{equation}
\label{eq:aug-speaker-representation}
    \operatorname{aug}_{s_{m+1}}
    = \alpha_{s_{m+1}} \cdot \mathbf{v}_{s_{m+1}}
    + \bigl(1-\alpha_{s_{m+1}}\bigr) \cdot \mathbf{b}_{s_{m+1}},
\end{equation}
where $\alpha_{s_{m+1}} \in [0,1]$ is an adaptive weight:
\begin{equation}
    \alpha_{s_{m+1}} = \min\left(1, \frac{|C_{s_{m+1}}|}{k}\right).
\end{equation}
A larger $\alpha_{s_{m+1}}$ indicates that $s_{m+1}$ has sufficient participation in $C_{-k}$, while a smaller $\alpha_{s_{m+1}}$ indicates sparse interaction history and thus places greater weight on $s_{m+1}$'s background information.

The augmented variants replace
$\max\;\operatorname{sim}(\cdot)$
with $\operatorname{sim}(e(c), \\ \operatorname{aug}_{s_{m+1}})$ in Eq.~(\ref{eq:ls-es-avg}) and Eq.~(\ref{eq:ls-es-max}).

\subsubsection{Local Speaker Topic-Based Alignment (LS-TA)}
To complement embedding-based similarity, we also consider topical alignment via topic modeling. Let $\theta(C_{-k}) \in \mathbb{R}^{L}$ denote the topic distribution over $L$ topics for the most recent $k$ turns, and let $\theta(C_{s_{m+1}}) \in \mathbb{R}^{L}$ denote the topic distribution of messages authored by the predicted speaker $s_{m+1}$ (obtained via LDA~\cite{blei2003latent} or BERTopic~\cite{grootendorst2022bertopic}). We quantify topic alignment using Jensen-Shannon divergence:
\begin{equation}
\label{eq:ls-ta}
    \text{LS-TA}(s_{m+1}, C_{-k}) = 1 - \sqrt{\text{JSD}(\theta(C_{s_{m+1}}) \| \theta(C_{-k}))}
\end{equation}
where $\text{JSD}(\cdot \| \cdot)$ measures the divergence between two probability distributions over topics. A lower divergence (higher score) indicates that the predicted speaker's historical messages are topically aligned with the recent discussion, suggesting relevant domain knowledge or engagement with the ongoing themes.

\subsection{Local Content Quality}\label{sec:local_content_measures}

Local content quality assess whether the predicted message content $u_{m+1}$ appropriately advances the conversation given its current state. Concretely, they capture behaviors such as responding to outstanding prompts, refining or extending partially formed ideas, proposing candidate next steps, reiterating prior content, or shifting to a new topic.

As indicated in Figure~\ref{fig:global-measurement-design}, conversations may or may not provide an explicit objective. Accordingly, we distinguish two scenarios for local content measures: \textbf{objective-guided} and \textbf{objective-free}.

In \textit{objective-guided} settings (Section~\ref{sec:local-content-supervised}), the conversation is equipped with an explicit objective $A$, typically instantiated as an agenda that can be decomposed into a finite set of discussion items. For example, in a paper-writing conversation, an agenda may include items such as ``finalize the introduction,'' ``revise the methodology section,'' and ``address reviewer comments on Figure~3.'' Content quality measures in this setting evaluate how effectively the predicted message $u_{m+1}$ advances progress toward $A$, e.g., by resolving pending items, moving to the next appropriate sub-goal, or providing missing information required to complete the task.

In \textit{objective-free} setting (Section~\ref{sec:local-content-unsupervised}), local content quality assesses whether $u_{m+1}$ forms a proper continuation of $C_{-k}$. We quantify this via a \textit{flow score} that characterizes how $u_{m+1}$ maintains topical coherence and manages local topic transitions---e.g., staying within the current thread, expanding to closely related subtopics, or smoothly shifting to another aspect of the discussion.


\subsubsection{Objective-guided}\label{sec:local-content-supervised}
In objective-guided settings, local content measures are defined relative to the conversation objective or, more specifically, the agenda $A$ (e.g., a meeting agenda, project milestones, or a troubleshooting checklist). Such an agenda can be viewed as a structured collection of content items that the conversation is expected to cover. While some agendas organize items as a linear sequence, others adopt richer hierarchical structures (e.g., a tree of subtopics under each major issue). For example, a project-planning discussion may organize items into phases such as ``design,'' ``implementation,'' and ``evaluation,'' each of which further decomposes into more fine-grained tasks.

We thus introduce an \emph{agenda-guided progression score (AP)} to measure how the predicted message content $u_{m+1}$ contributes with respect to the current agenda state, either by enriching the discussion of the active item or by supporting a transition toward subsequent items that have received less coverage.

\paragraph{Agenda Representation} Given the agenda $A$ of a conversation, we assume that $A$ is instantiated as a finite set of agenda items $\{a_1,\ldots,a_n\}$. We represent structural relations among these items using a directed graph $G=(V,E)$, where $V=\{a_1,\ldots,a_n\}$ and $(a_i,a_j)\in E$ indicates that $a_j$ is a follow-up item of $a_i$.

\paragraph{Agenda State Tracking} We track which agenda item is most relevant to the discussion in $C_{-k}$ and assess if this item has received sufficient coverage in $C_{-k}$, i.e., has been discussed with enough information or diverse perspectives. We formalize this notion with the information coverage score (InfoCov) illustrated below.

For each agenda item $a_i \in \mathcal{A}=\{a_1,\ldots,a_n\}$, we first collect the turns in $C_{-k}$ that are relevant to $a_i$:
\[
C_{a_i} = \{\,c \in C_{-k} \mid \operatorname{sim}(e(a_i), e(c)) > \tau_{\text{rel}}\,\},
\]
where $\tau_{\text{rel}}$ is a relevance threshold (e.g., $0.6$). We then select $a_{i^*}$ as the agenda item most relevant (i.e., most heavily discussed) to $C_{-k}$:
\[
i^* = \arg\max_{i \in \{1,\ldots,n\}} |C_{a_i}|.
\]

Given the selected item $a_{i^*}$, we measure its information coverage by combining (i) how many relevant turns it receives and (ii) how semantically diverse those turns are. Let $n_{i^*}=|C_{a_{i^*}}|$. We define
\begin{equation}
\label{eq:infocov}
\begin{split}
\text{InfoCov}(a_{i^*}, C_{-k})
&= \bigl(1 - e^{-\frac{n_{i^*}}{\gamma}}\bigr)\cdot
\frac{1}{\binom{n_{i^*}}{2}} \\
&\quad \cdot \sum_{\substack{c, c' \in C_{a_{i^*}}\\ c \neq c'}}
\Bigl(1 - \operatorname{sim}(e(c), e(c'))\Bigr).
\end{split}
\end{equation}

where $\gamma$ is a scale parameter. The factor $\bigl(1-e^{-n_{i^*}/\gamma}\bigr)$ increases with $n_{i^*}$, and the second term is the average pairwise dissimilarity among relevant turns; thus, $\text{InfoCov}(a_{i^*},C_{-k})$ increases only when $a_{i^*}$ is discussed by a sufficient number of semantically non-redundant messages. We say that the current item $a_{i^*}$ is \emph{saturated} if $\text{InfoCov}(a_{i^*},C_{-k})>\tau_{\operatorname{cov}}$ for a coverage threshold $\tau_{\operatorname{cov}}$.


\paragraph{Agenda-Guided Progression Score} Given the predicted message $u_{m+1}$ and the current agenda item $a_{i^*}$, we distinguish two cases depending on whether the current agenda $a_{i^*}$ is saturated according to $\text{InfoCov}(a_{i^*}, C_{-k})$.

\textbf{Case 1: Current agenda item is not saturated}. 
Intuitively, $a_{i^*}$ has not yet been covered by a sufficiently diverse set of messages, so further discussion may still add value. We measure whether $u_{m+1}$ enriches $a_{i^*}$ by contributing to its information diversity:
\begin{equation}
\text{InfoGain}(u_{m+1}, a_{i^*}) =
\frac{1}{n_{i^*}}
\sum_{c \in C_{a_{i^*}}}
\bigl(1 - \text{sim}(e(u_{m+1}), e(c))\bigr),
\end{equation}
The agenda-guided progression score is then
\begin{equation}
\text{AP}(u_{m+1}, a_{i^*}) =
\text{sim}(e(u_{m+1}), e(a_{i^*})) \cdot \text{InfoGain}(u_{m+1}, a_{i^*}),
\end{equation}
which assigns higher scores to messages that both are relevant to $a_{i^*}$ and introduce content that is semantically distinct from previous contributions on $a_{i^*}$.

\begin{algorithm}[t]
\caption{Next agenda item selection via DFS}
\label{alg:next-agenda-item}
\begin{algorithmic}[1]
\Procedure{NextAgendaItem}{$G, i^*$}
    \State Mark all items in $G$ as undiscovered
    \State \Return \Call{DFS}{$G, i^*$}
\EndProcedure

\Procedure{DFS}{$G, i$}
    \State \Return \textsc{None} \textbf{if} $i$ is discovered
    \State Mark $i$ as discovered
    \State \Return $i$ \textbf{if} $\text{InfoCov}(a_i, C_{-k}) \le \tau_{\operatorname{cov}}$
    
    \ForAll{directed edges $(i, j)$ in $G$} \Comment{successors of $a_i$}
        \State \Return $r \gets \Call{DFS}{G, j}$ \textbf{if} $r \neq \textsc{None}$
    \EndFor
    
    \ForAll{directed edges $(j, i)$ in $G$} \Comment{predecessors of $a_i$}
        \State \Return $r \gets \Call{DFS}{G, j}$ \textbf{if} $r \neq \textsc{None}$
    \EndFor
    \State \Return \textsc{None}
\EndProcedure
\end{algorithmic}
\end{algorithm}

\textbf{Case 2: Current agenda item is saturated}. When the current agenda item has been discussed sufficiently, we look for less-covered items elsewhere in the agenda graph $G$. To obtain the next agenda item, Algorithm~\ref{alg:next-agenda-item} performs a depth-first search (DFS) starting from $a_{i^*}$ and finds the first unsaturated item. Let $a_{i_{\text{next}}}$ denote the next agenda item found, the agenda-guided progression score measures whether $u_{m+1}$ enriches $a_{i_{\text{next}}}$:
\begin{equation*}
\text{AP}(u_{m+1}, a_{i_{\text{next}}}) =
\text{sim}(e(u_{m+1}), e(a_{i_{\text{next}}})) \cdot \text{InfoGain}(u_{m+1}, a_{i_{\text{next}}}).
\end{equation*}
This rewards messages that advance the conversation to items requiring further exploration.

\subsubsection{Objective-free}\label{sec:local-content-unsupervised}
In objective-free conversations, a critical aspect of evaluating the quality of the predicted next message is its \textbf{relevance} to the conversation history. More specifically, we assess whether the predicted message content \textbf{flows} naturally with respect to prior turns. Good flow means the message is logically connected to some aspect of the preceding discussion---advancing the conversation coherently rather than reiterating previously covered points or introducing disconnected topics that break conversational continuity. Evaluating content flow aims to ensure that the predicted message makes sense within the conversational trajectory.

In the following, we present multiple quantitative metrics for assessing content flow score.

\paragraph{Lexical Novelty Rate (LNR)}
One way to quantify local conversational flow is to measure how many \emph{new} lexical units the predicted message introduces relative to the recent context. We extract a set of lexical units from an utterance, including lemmatized content words, named entities, and content-word bigrams. Let $E(u)$ denote the set of extracted units from utterance $u$, and let $E(C_{-k})=\bigcup_{(s,u)\in C_{-k}} E(u)$. The set of novel units in $u_{m+1}$ is
\[
E_{\text{novel}}(u_{m+1},C_{-k})=E(u_{m+1}) \setminus E(C_{-k}).
\]
The lexical novelty rate is defined as
\begin{equation}
\label{eq:lnr}
\text{LNR}(u_{m+1},C_{-k})=\frac{|E_{\text{novel}}(u_{m+1},C_{-k})|}{|E(u_{m+1})|}.
\end{equation}
Extreme values typically indicate poor flow for different reasons: $\text{LNR}\approx 0$ suggests heavy lexical reuse (repetition), while $\text{LNR}\approx 1$ suggests largely unseen vocabulary (topic drift). Coherent continuations tend to yield intermediate values, introducing some new units while maintaining lexical overlap with $C_{-k}$.

In many cases, surface-level lexical novelty may over-count paraphrases (e.g., synonyms). To reduce this effect, we treat a unit as \emph{truly novel} only if it is not semantically close to any unit in the recent context. Let $e(w)$ be the embedding of unit $w$, We consider $w$ truly novel if its maximum embedding similarity to any unit in $E(C_{-k})$ falls below a threshold $\tau_{\text{LNR}}$:
\[
\begin{split}
E_{\text{tn}}(u_{m+1},C_{-k})
=\Bigl\{\, w\in E(u_{m+1}) \ \Big|\ 
\max_{w_c\in E(C_{-k})} \\ \operatorname{sim}\bigl(e(w),e(w_c)\bigr)
< \tau_{\text{LNR}} \Bigr\}.
\end{split}
\]
The embedding-aware lexical novelty rates are:
\begin{equation}
\label{eq:lnr-e}
\text{LNR-E}(u_{m+1},C_{-k})=\frac{|E_{\text{tn}}(u_{m+1},C_{-k})|}{|E(u_{m+1})|},
\end{equation}
and an IDF-weighted (inverse document frequency) variant,
\begin{equation}
\label{eq:lnr-e-w}
\text{LNR-E-w}(u_{m+1},C_{-k})
=\frac{\sum_{w\in E_{\text{tn}}(u_{m+1},C_{-k})}\text{IDF}(w)}
{\sum_{w\in E(u_{m+1})}\text{IDF}(w)}.
\end{equation}
As with LNR, extreme values of LNR-E/LNR-E-w indicate either repetition (near $0$) or topical drift (near $1$), while intermediate values correspond to stronger flow with the immediate context.

\paragraph{Message-Level Semantic Novelty Score (M-SNS)}
\nop{
Lexical novelty may miss message-level semantically novel content. We measure message-level semantic novelty score (M-SNS) as the distance from $u_p$ to its nearest neighbor in the recent conversation history, denoted as M-SNS-min:
\begin{equation}
\text{M-SNS-min} = \min_{c \in C_{-k}} \left(1 - \text{sim}(\mathbf{e}(u_p), \mathbf{e}(c))\right)
\end{equation}
where $\mathbf{e}(\cdot)$ denotes message embeddings and $\text{sim}(\cdot, \cdot)$ is cosine similarity. A very high flowing score in this case indicates that $u_p$'s semantic content is distant from all recent messages, which means it is very likely to be off-topic irrelevant content, while a very low score means the content is nearly identical to existing messages, suggesting verbatim reiteration with no novelty. Instead, a \textbf{moderate score} indicates good flow, where the predicted message maintains semantic coherence with the conversation history while introducing novel content.

Alternatively, we can measure the average distance to capture overall distinctiveness, denoted as M-SNS-avg:
\begin{equation}
\text{M-SNS-avg} = \frac{1}{|C_{-k}|} \sum_{c \in C_{-k}} \left(1 - \text{sim}(\mathbf{e}(u_p), \mathbf{e}(c))\right)
\end{equation}

The minimum distance variant captures the most similar precedent (``Is there anything like this?''), while the average variant captures overall distinctiveness from the recent discussion.
}

Lexical novelty is a surface-level signal and may miss semantically novel content at the message level. We therefore measure the \emph{message-level semantic novelty score (M-SNS)} as the embedding distance between the predicted message $u_{m+1}$ and recent turns in $C_{-k}$. 

We first define the average-distance variant (M-SNS-avg), which measures overall semantic distinctiveness from the recent context:
\begin{equation}
\text{M-SNS-avg}(u_{m+1}, C_{-k})
= \underset{c \in C_{-k}}{\mathrm{avg}} \left(1 - \operatorname{sim}\bigl(e(u_{m+1}), e(c)\bigr)\right).
\end{equation}
A high score indicates that $u_{m+1}$ is broadly distant from the recent discussion (potential topic drift), while a very low score indicates heavy semantic overlap with the context (reiteration). Intermediate scores typically indicate that $u_{m+1}$ remains connected to $C_{-k}$ while introducing new information.

We also consider a nearest-neighbor variant (M-SNS-min), which measures the distance from $u_{m+1}$ to its most similar precedent in $C_{-k}$:
\begin{equation}
\text{M-SNS-min}(u_{m+1}, C_{-k})
= \min_{c \in C_{-k}} \left(1 - \operatorname{sim}\bigl(e(u_{m+1}), e(c)\bigr)\right).
\end{equation}
Compared to M-SNS-avg, M-SNS-min is sensitive to whether $u_{m+1}$ closely matches \emph{any} recent turn (i.e., ``is there anything like this?''), whereas M-SNS-avg reflects the \emph{overall} distinctiveness from $C_{-k}$.

\paragraph{Dialogue-Act Transition (DAF)}
A dialogue act labels the conversational \emph{function} of a message (e.g., question, answer, acknowledgment, request, prompt). Many continuations are predictable at this functional level: certain acts tend to follow others with higher probability over short windows (e.g., questions are frequently followed by answers). The \emph{Dialogue-Act Transition Fit} measures whether the act assigned to the predicted message $u_{m+1}$ is a plausible next act given the recent act sequence in $C_{-k}$.

The assumption underlying this metric is that a plausible dialogue act (e.g., answering a question raised in the recent context) is more likely to exhibit good local content flow.

Let $D(u)$ denote the dialogue act assigned to message $u$. Given $C_{-k}$, let $\mathbf{d}_{-k}=\langle D(u_{m-k+1}),\ldots,D(u_m)\rangle$
denote the  recent act sequence in $C_{-k}$. We model the conditional probability $p(d \mid \mathbf{d}_{-k})$. Given $\hat{d}=D(u_{m+1})$, the dialogue-act transition fit is
\begin{equation}
\label{eq:daf}
\text{DAF}(u_{m+1}, C_{-k}) = p(\hat{d} \mid \mathbf{d}_{-k}).
\end{equation}
A higher DAF value indicates that the predicted dialogue act of $u_{m+1}$ is more consistent with the local act dynamics in $C_{-k}$.

For message-level act tagging, existing classifiers such as BERT-based encoders~\cite{zhou2025comprehensive} are widely used. To model act dynamics, several approaches can be employed, including Hidden Markov Models (HMMs)~\cite{rabiner2003introduction} or transformer-based sequence models~\cite{han2021transformer}.

\paragraph{Log-Likelihood-Based Method (LL)}
The log-likelihood-based measure evaluates how \emph{expected} the predicted message content $u_{m+1}$ is given $C_{-k}$. Intuitively, if a language model were to generate the next message conditioned on $C_{-k}$, would it assign high probability to $u_{m+1}$? Autoregressive language models are suitable for this estimation because they define token-level conditional probabilities; aggregating these probabilities over the tokens in $u_{m+1}$ yields a contextual fit score.

The assumption underlying this metric is that content assigned higher likelihood by the language model is more likely to exhibit strong local flow.

Let $u_{m+1}=\langle u_{m+1,1},\ldots,u_{m+1,L_{m+1}}\rangle$ denote the token sequence of the predicted message, where $L_{m+1} = |u_{m+1}|$ is the sequence length. A language model provides conditional token probabilities $p(u_{m+1,j}\mid u_{m+1,<j},C_{-k})$. We compute the score as the exponentiated average token log-likelihood:
\begin{equation}
\label{eq:ll}
\text{LL}(u_{m+1}, C_{-k}) = \exp\left(\frac{1}{L_{m+1}}\sum_{j=1}^{L_{m+1}}\log p(u_{m+1,j} \mid u_{m+1,<j}, C_{-k})\right).
\end{equation}
Higher values indicate that $u_{m+1}$ is a more plausible continuation of $C_{-k}$ under the language model's learned distribution.

\paragraph{Topic Expansion Score (TES)}\label{sec:content_coherence_topic}
A predicted message with good flow typically preserves the established themes in the recent context while introducing moderate novelty---avoiding both topical disconnection and verbatim reiteration. Topic models provide a higher-level view of this trade-off by encoding text as a distribution over latent topics~\cite{vayansky2020review}. However, directly comparing $\theta(C_{-k})$ with $\theta(u_{m+1})$ is often uninformative because $u_{m+1}$ can be short or generic, and its topic mixture may not reliably reflect the ongoing themes without local context. We therefore (i) select a context window whose topic mixture is stable, and (ii) evaluate the topical contribution of $u_{m+1}$ using a small augmentation window to avoid dilution by a long context.

Formally, let $\theta(X)\in\mathbb{R}^{L}$ denote the topic distribution over $L$ topics for any text $X$, and let $\mathrm{JSD}(\cdot,\cdot)$ denote Jensen-Shannon divergence. We use the transformed divergence $1-\sqrt{\mathrm{JSD}(\cdot,\cdot)}\in[0,1]$ as a similarity score.

\textbf{Step 1: choose a stable context size $h^\ast$.}
Starting from $h\in\{\Delta,2\Delta,3\Delta,\ldots\}$, we expand the context in increments of $\Delta$ and stop if adding $\Delta$ more turns would substantially shift the topic mixture:
\[
h^\ast=\min\Bigl\{h \mid \mathrm{JSD}\bigl(\theta(C_{-h}),\theta(C_{-(h+\Delta)})\bigr)>\tau_{\text{topic}}\Bigr\}.
\]
We then set $\pi^\ast=\theta(C_{-h^\ast})$ as the established topic mixture.

\textbf{Step 2: compute the topic expansion score.}
To reflect the topic mixture of the predicted message $u_{m+1}$, we augment $u_{m+1}$ with a small window of previous messages, $C_{-\ell}$ where $\ell \ll h^\ast$ (e.g., $\ell\in\{2,3\}$), as supporting context, and obtain its topic mixture as
\[
Q_{m+1}=\theta(C_{-\ell}\oplus u_{m+1}).
\]
Let $\mathcal{T}_{\text{dom}}(\pi^\ast,\rho)$ be the set of dominant topics that together account for a fraction $\rho$ of the mass in $\pi^\ast$ (e.g., $\rho=0.8$). We define the \emph{topic expansion score} as the probability mass that $Q_{m+1}$ assigns to non-dominant topics:
\begin{equation}
\label{eq:tes}
\text{TES}(u_{m+1}, C_{-k^\ast}, \rho)
= \sum_{l \notin \mathcal{T}_{\text{dom}}(\pi^\ast,\rho)} Q_{m+1, l}.
\end{equation}
Intuitively, TES measures how much the predicted message shifts probability mass toward topics that are not central in the established context.
$\text{TES}\approx 0$ indicates little topical expansion (potential redundancy), whereas $\text{TES}\approx 1$ indicates strong expansion into previously inactive themes (potential disconnection). Intermediate values suggests $u_{m+1}$ remains topically connected while introducing new information.

\subsection{Local speaker--content Consistency}
\label{sec:local_speaker_content_consistency}

Local speaker--content consistency measures assess whether the predicted message content $u_{m+1}$ is characteristic of the predicted speaker $s_{m+1}$, based on the speaker's established communication patterns and domain expertise. While local speaker measures evaluate whether $s_{m+1}$ is an appropriate next speaker and local content measures evaluate whether $u_{m+1}$ is an appropriate continuation, speaker--content consistency bridges the two by asking: \emph{does this content sound like something this speaker would say?}

This measure is particularly useful for detecting attribution errors in multi-party conversations, where a message may be contextually appropriate yet inconsistent with the attributed speaker's role, expertise, or style.

Formally, speaker--content consistency quantifies the alignment between $u_{m+1}$ and the historical messages from $s_{m+1}$ (and, when available, the background information $B_{s_{m+1}}$). A higher consistency score indicates that $u_{m+1}$ aligns thematically or stylistically with how $s_{m+1}$ has communicated, whereas a lower score suggests a potential speaker--content mismatch.
To this, we leverage embedding-based metrics to quantify local speaker--content consistency.

\subsubsection{Embedding-based Local speaker--content Consistency Measures}

We measure the consistency between the predicted content $u_{m+1}$ and the predicted speaker $s_{m+1}$ by comparing the embedding of $u_{m+1}$ with embeddings derived from $s_{m+1}$'s established communication patterns.

\paragraph{Local speaker--content Consistency via Average Embedding Similarity (LSCC-ES-avg)}
It measures the average similarity between $u_{m+1}$ and all previous utterances by $s_{m+1}$ in the recent context $C_{-k}$:
\begin{equation}
\label{eq:lscc-es-avg}
    \text{LSCC-ES-avg}(u_{m+1}, s_{m+1}, C_{-k}) = \operatorname*{avg}_{c \in C_{s_{m+1}}} \text{sim}(e(u_{m+1}), e(c)),
\end{equation}
where $C_{s_{m+1}} = \{(s_i, u_i) \in C_{-k} \mid s_i = s_{m+1}\}$ denotes the subset of messages in $C_{-k}$ authored by the predicted speaker.

This metric captures overall conversational consistency: if $u_{m+1}$ discusses content similar to what $s_{m+1}$ typically discusses, the average similarity will be high.

\paragraph{Local speaker--content Consistency via Maximum Embedding Similarity (LSCC-ES-max)}
The maximum consistency score identifies the strongest topical connection between $u_{m+1}$ and any single previous message from $s_{m+1}$:
\begin{equation}
\label{eq:lscc-es-max}
    \text{LSCC-ES-max}(u_{m+1}, s_{m+1}, C_{-k}) = \max_{c \in C_{s_{m+1}}} \text{sim}(e(u_{m+1}), e(c)).
\end{equation}

This metric is useful for detecting whether $u_{m+1}$ continues or extends any specific thread that $s_{m+1}$ previously initiated, even if their overall participation spans diverse topics.

\paragraph{Local speaker--content Consistency via Minimum Embedding Similarity (LSCC-ES-min)}
The minimum consistency score captures the weakest link, identifying whether $u_{m+1}$ has at least some connection to $s_{m+1}$'s previous contributions:
\begin{equation}
\label{eq:lscc-es-min}
    \text{LSCC-ES-min}(u_{m+1}, s_{m+1}, C_{-k}) = \min_{c \in C_{s_{m+1}}} \text{sim}(e(u_{m+1}), e(c)).
\end{equation}

A high minimum score indicates that $u_{m+1}$ is broadly consistent with $s_{m+1}$'s entire communication history in the context, rather than only aligning with a subset of their utterances.

\paragraph{Incorporating Background Information for Sparse Conversation History} 
Similar to the embedding-based local speaker measures (Section~\ref{sec:local-speaker-embedding}), when $|C_{s_{m+1}}|$ is small, the consistency scores above may be unreliable or undefined. However, $s_{m+1}$ might still be the appropriate speaker if their background information $B_{s_{m+1}}$ (e.g., job title, expertise areas, profile description), if given, aligns with the predicted content.



%
To address this, we leverage the augmented speaker representation established in Eq.~(\ref{eq:aug-speaker-representation}). The augmented speaker--content consistency score (LSCC-ES-aug) is then:
\begin{equation}
\label{eq:lscc-es-aug}
    \text{LSCC-ES-aug}(u_{m+1}, s_{m+1}, C_{-k})
    = \operatorname{sim}\!\left(e(u_{m+1}), \operatorname{aug}_{s_{m+1}}\right).
\end{equation}

This formulation ensures that even when $s_{m+1}$ has minimal conversation history (e.g., a domain expert who has been silent), the consistency score can still reflect alignment between the predicted content and the speaker's documented expertise.

\section{Detailed Global Evaluation}
\label{sec:global-old}

Global evaluation assesses the quality of an \emph{entire multi-party conversation}. We represent a conversation as a sequence of turns, where each turn is a speaker-utterance pair. When the conversation history is available, let
$C=\langle (s_1,u_1),(s_2,u_2),\ldots,(s_m,u_m)\rangle$
denote the history, and let
$\tilde{C}=\langle (\tilde{s}_{m+1},\tilde{u}_{m+1}),\ldots,(\tilde{s}_{m+T},\tilde{u}_{m+T})\rangle$
denote the model-generated continuation of $T$ turns.
In this case, the evaluation object is the aggregated conversation trajectory
$\hat{C} = C \oplus \tilde{C}$.
In cases without given conversation history, in other words, the conversation is generated from-scratch by the model (e.g., conditioned on a given topic, scenario, or initial seed), we simply set
$\hat{C}=\langle (\tilde{s}_{1},\tilde{u}_{1}),\ldots,(\tilde{s}_{T},\tilde{u}_{T})\rangle$.
In both settings, global evaluation focuses on \emph{long-horizon properties} when assessing the conversation as a whole---e.g., how participation is distributed across speakers, whether information is collaboratively constructed or centralized, and whether the discussion makes steady progress over time (toward an explicit objective when available).

At the global level, the three measurement aspects introduced in Section~\ref{sec:benchmark-objective} specialize as follows:
\begin{enumerate}
    \item \textbf{Global Speaker Modeling} (Appendix~\ref{sec:global_speaker_measures}) evaluates whether the \emph{speaker assignments} across $\hat{C}$ exhibit plausible multi-party interaction patterns, including participation balance and the distribution of contributions across speakers.
    \item \textbf{Global Content Quality} (Appendix~\ref{sec:global_content_measures}) evaluates \emph{what is said} across $\hat{C}$, emphasizing long-range progression and structural coherence (and, when objectives are provided, task/agenda completion and progression efficiency).
    \item \textbf{Global speaker--content Consistency} (Appendix~\ref{sec:global_speaker_content_consistency}) evaluates whether each utterance remains consistent with its attributed speaker throughout $\hat{C}$, allowing speakers to exhibit one or multiple stable characteristics (e.g., topic, style, role) while penalizing inconsistent drift.
\end{enumerate}

\subsection{Global Speaker Modeling}
\label{sec:global_speaker_measures}

One straightforward approach to global speaker modeling is to accumulate the local speaker modeling scores introduced in Section~\ref{sec:local_speaker_measures}. Concretely, we view generation of an entire conversation as an iterative next-turn prediction process. At step $t$, the model conditions on the available context---the history $C$ (if provided) concatenated with the previously generated turns---and predicts the next speaker. We score each predicted speaker at every step and report aggregate statistics (e.g., mean accuracy) over the full roll-out.

However, this approach has two notable limitations. First, it can be computationally expensive for long horizons conversations. Second, while local speaker modeling evaluates whether \emph{each} predicted speaker is correct, simple aggregation may fail to capture whether the resulting interaction exhibits a reasonable \emph{participation pattern}. That is, a locally plausible next speaker prediction does not necessary lead to a globally plausible turn-taking pattern. For example, a next-speaker prediction model can simply select from a small set of dominant speakers (based on its high participation frequency (PF) and relevance to ongoing discussions (LS-ES)), but overall, it may yield uneven or partial participation.

Motivated by these issues, we complement global speaker accuracy with measures of \emph{global speaker diversity and dynamics} in the subsequent section.

\subsubsection{Global Speaker Diversity and Dynamics}

We evaluate global speaker dynamics from two complementary perspectives that characterize group interaction patterns. \textit{Structural diversity} measures how evenly speakers participate, detecting whether $\hat{C}$ is dominated by a single speaker or broadly distributed. \textit{Semantic concentration} measures how semantic and information coverage is allocated across speakers, identifying whether the discussion is driven by a central authority or collaboratively constructed.

\paragraph{Structural Diversity.}
We first measure the evenness of participation. Given a set of $S$ total speakers in $\hat{C}$. We compute the empirical turn distribution $P=\{p_1,\ldots,p_S\}$, where $p_i$ is the fraction of turns attributed to speaker $s_i$. We then define the normalized speaker entropy (NSE) as:
\begin{equation}
\label{eq:NSE}
\text{NSE}(\hat{C})=\frac{-\sum_{i=1}^{S} p_i \log_2 p_i}{\log_2 S}.
\end{equation}
A score near $0$ indicates highly imbalanced participation (e.g., a near-monologue), while a score near $1$ indicates near-uniform participation across speakers.

\paragraph{Semantic Concentration.}
To characterize how the overall semantics and the information of the conversation is distributed across speakers, we measure each speaker's contribution to the conversation's overall information spread.
We compute the spread $R(\hat{C})$ as the average minimum spanning tree (MST) edge length over message embeddings under the distance
$d(u,u') = 1 - \text{sim}(e(u),e(u'))$.
For each speaker $s_i$, we quantify their contribution via a leave-one-speaker-out comparison:
\[
\Gamma_{s_i} = \max\{0,\, R(\hat{C}) - R(\hat{C} \setminus \hat{C}_{s_i})\},
\]
where $\hat{C}_{s_i}$ denotes the set of turns in $\hat{C}$ attributed to speaker $s_i$. Intuitively, $\Gamma_{s_i}$ is large when removing $s_i$ noticeably reduces the information spread, indicating that $s_i$ contributed unique content or ideas; it is small when $s_i$'s messages are largely redundant with the conversation.

We then apply the Gini coefficient to quantify how concentrated these contributions are across speakers:
\[
\text{SC-Gini}(\hat{C}) =
\frac{\sum_{i=1}^{S}\sum_{j=1}^{S}\lvert \Gamma_{s_i}-\Gamma_{s_j}\rvert}
{2S \sum_{i=1}^{S}\Gamma_{s_i}}.
\]
A higher SC-Gini indicates information centralization, where a small number of speakers account for most contribution. A lower SC-Gini indicates more collaborative construction, where the overall information and semantics are more evenly distributed.

\subsection{Global Content Quality}
\label{sec:global_content_measures}

Global content quality evaluates whether the conversation $\hat{C}$ exhibits coherence over long-horizon---i.e., the interaction introduces substantive content, maintains a sensible conversational trajectory, and makes progress over time. As in Figure~\ref{fig:global-measurement-design}, we distinguish two settings depending on whether an explicit objective is provided. In \textit{objective-guided} generation, the conversation comes with an objective $\mathcal{A}$ (often instantiated as a collection of agenda items), and global content measures quantify end-to-end success and progression toward $A$ (e.g., task success, agenda completion, and progression efficiency). In \textit{objective-free} generation, there is no explicit $\mathcal{A}$; global content quality measures the overall semantic progression of $\hat{C}$, capturing whether the discussion evolves in a coherent, non-trivial manner (e.g., sustained development vs.\ stagnation or uncontrolled topic drift).

\subsubsection{Objective-guided} 
\nop{
Global supervised content accuracy evaluates the quality of content generation across the entire conversation trajectory. Given a conversation history $C$ up to turn $m$, a generation model produces a complete conversation:
\[
\hat{C} = \langle (\hat{s}_{m+1}, \hat{u}_{m+1}), \dots, (\hat{s}_{n}, \hat{u}_{n}) \rangle,
\]
where each turn $(\hat{s}_{m+i}, \hat{u}_{m+i})$ is conditioned on $C$ and all previously generated turns. 

A straightforward approach would be to aggregate local supervised content accuracy scores across all generated turns. However, this often leads to a model generating a sequence of locally plausible messages---each exhibiting high agenda adherence in isolation---that collectively fail to fulfill the overall conversational objective or comprehensively address the entire agenda. To capture this divergence between local optimality and global utility, we measure not only if the conversation moves forward message-by-message, but if it effectively and logically converges to a successful terminal state from four aspects:
}

Global \emph{objective-guided} content quality evaluates the quality of content generation over the full conversation trajectory. A straightforward approach is to aggregate local objective-guided content quality measures across generated turns. However, a model may produce a sequence of locally plausible, agenda-aligned messages that nevertheless fails to satisfy the overall objective. To capture global utility, we evaluate whether $\hat{C}$ not only advances step-by-step, but also \emph{effectively and comprehensively} covers the agenda and reaches the end goal. Concretely, we assess this from four complementary aspects:

\paragraph{Task Success ($\Phi(\cdot)$)}
We view $\hat{C}$ as a procedural transformation from an initial intent to a final outcome $O_{\hat{C}}$---either a tangible artifact (e.g., a filled form or a code snippet) or an abstract terminal state (e.g., reaching agreement). Under this view, task success can be evaluated by a validation function (e.g., unit tests) that checks whether the outcome satisfies a set of acceptance criteria.

We define a binary task-success score $\Phi(\hat{C})\in\{0,1\}$ via two cases, depending on whether the outcome is explicit or implicit.

\textbf{Case 1: Artifact Validation (Explicit Output).} 
If the conversation goal is to produce an explicit artifact $O_{\hat{C}}$ (e.g., a code snippet), we validate it against a criteria set $\mathcal{K}=\{\kappa_1,\ldots,\kappa_m\}$, where each predicate $\kappa_j: O_{\hat{C}}\rightarrow\{0,1\}$ encodes a specific requirement (e.g., ``is valid JSON'', ``all mandatory fields present''). Success requires the satisfaction of all constraints:
\begin{equation}
\label{eq:isvalid}
\text{isValid}(O_{\hat{C}}, \mathcal{K}) \;=\; \prod_{j=1}^{m} \kappa_j(O_{\hat{C}}).
\end{equation}
Thus, for conversations explicitly requiring output artifacts, $\Phi(\hat{C}) = \text{isValid}(O_{\hat{C}}, \mathcal{K})$.

\textbf{Case 2: State Validation (Implicit Output).} 
If the outcome is an abstract terminal state (e.g., ``Knowledge Transfer'' or ``Agreement''), we treat the agenda's terminal item $a_{\text{end}}$ as the required ``return condition.'' Success is achieved if $a_{\text{end}}$ is saturated. We define the saturation indicator for any agenda item $a$ as:
\begin{equation}
\label{eq:saturation_global}
\mathrm{Sat}(a,\hat{C}) \;=\; \mathbb{I}\!\left(\text{InfoCov}(a,\hat{C}) > \tau_{\text{cov}}\right),
\end{equation}
where $\text{InfoCov}(a,\hat{C})$ measures how well the content of $\hat{C}$ semantically covers the agenda item $a$---operationalized in Eq.~(\ref{eq:infocov}) as an embedding-based coverage score between $a$ and the conversation content---and $\tau_{\text{cov}}$ is a sufficiency threshold. Therefore, for conversations that end with an intended state: $\Phi(\hat{C}) = \mathrm{Sat}(a_{\text{end}}, \hat{C})$.

A high score of $\Phi(\hat{C})$ in either cases indicates the conversation successfully produced a valid output that met all constraints (e.g., a compiling code snippet or a confirmed appointment). A low score indicates the process terminated with an invalid output (e.g., a syntax error in generated code) or did not fulfill the terminal state.

\paragraph{Agenda Completion Rate (ACR)} 
While $\Phi(\hat{C})$ captures whether the conversation reaches a successful terminal outcome, it does not reveal \emph{how comprehensively} the discussion covers the intended agenda. To evaluate completeness, we quantify the fraction of agenda items $\mathcal{A}=\{a_{1},\ldots,a_{n}\}$ that are sufficiently addressed in $\hat{C}$. Using the saturation indicator $\mathrm{Sat}(a,\hat{C})$ defined in Eq.~(\ref{eq:saturation_global}), the agenda completion rate is:
\begin{equation}
\label{eq:acr}
\text{ACR}(\hat{C},\mathcal{A}) \;=\; \frac{1}{n}\sum_{i=1}^{n} \mathrm{Sat}(a_i,\hat{C}).
\end{equation}
    
A higher ACR indicates a more comprehensive discussion that covers a larger portion of the agenda items; whereas a low ACR implies an incomplete discussion with omissions, often mimicking a rushed discussion with many shortcuts. 

\paragraph{Progression Efficiency (PE)} 
ACR measures \emph{how much} of the agenda is covered but does not capture \emph{how efficiently} the conversation reaches that coverage. In practice, two conversations can achieve the same ACR while differing substantially in length: one may cover agenda items with concise, purposeful exchanges, whereas another may require redundant turns or repeated clarification. Progression efficiency therefore measures the \emph{cost} of achieving coverage---how many turns are used to saturate the agenda items.

Let $\mathcal{A}_{\text{sat}}=\{a\in\mathcal{A}\mid \mathrm{Sat}(a,\hat{C})=1\}$ be the set of saturated items (Eq.~\ref{eq:saturation_global}). For each $a\in\mathcal{A}_{\text{sat}}$, let $\hat{C}^{(a)}$ be the subset of messages in $\hat{C}$ that are relevant to $a$, where a message $u$ is considered relevant if $\text{sim}(e(u),e(a))>\tau_{\text{rel}}$ (as in Sec.~\ref{sec:local-content-supervised}). Progression efficiency is the average number of relevant turns per saturated item:
\begin{equation}
\label{eq:pe}
\text{PE}(\hat{C},\mathcal{A}) \;=\; \frac{1}{|\mathcal{A}_{\text{sat}}|+\epsilon}\sum_{a\in\mathcal{A}_{\text{sat}}} \bigl|\hat{C}^{(a)}\bigr|.
\end{equation}
    
A lower PE indicates higher efficiency, where items are saturated with fewer messages. A higher PE indicates lower efficiency, suggesting more extensive discussion per agenda item.

\paragraph{Conversational Structure (CS)}
Beyond \emph{whether} agenda items are eventually covered (ACR) and \emph{how efficiently} they are addressed (PE), we also evaluate \emph{how} the conversation progresses: does $\hat{C}$ follow an agenda-guided trajectory, or does it irregularly jump between topics? Conversational Structure (CS) quantifies the logical regularity of agenda transitions of $\hat{C}$. In other words, CS measures whether the sequence of agenda items expressed in $\hat{C}$ progresses largely in the same order as the reference agenda order $\mathcal{L}$. Specifically, CS is computed per the following steps. 

\textbf{Step 1 (Reference order via agenda linearization).}
Since agendas may be non-linear (e.g., an acyclic graph of dependent agenda items), we first linearize the agenda $\mathcal{A}$ into an ordered reference list $\mathcal{L}$ using a DFS-based traversal procedure (Alg.~\ref{alg:linearization}). The list $\mathcal{L}$ serves as a canonical ``expected'' progression order used to assess whether the conversation follows a natural flow through agenda items.

\begin{algorithm}
\caption{Agenda Linearization}
\label{alg:linearization}
\begin{algorithmic}[1]
\Procedure{Linearize}{Graph $G$ of agenda $\mathcal{A}$, start item $a_{\text{start}}$}
    \State $\mathcal{L} \gets$ empty ordered list
    \State \Call{Traverse}{$a_{\text{start}}$}
    \State \Return $\mathcal{L}$
\EndProcedure
\Procedure{Traverse}{$a_i$}
    \If{$a_i \in \mathcal{L}$}
        \State \Return
    \EndIf
    \State Append $a_i$ to $\mathcal{L}$
    \For{each child $a_j$ of $a_i$ in $G$}
        \State \Call{Traverse}{$a_j$}
    \EndFor
\EndProcedure
\end{algorithmic}
\end{algorithm}

\textbf{Step 2 (Agenda trajectory extraction).}
We next map the generated conversation $\hat{C}$ to an agenda trajectory $\hat{Z}$. We partition $\hat{C}$ into overlapping blocks $B_1,B_2,\ldots$ of size $w$ with step size $\delta$ (e.g., $w=5,\delta=2$). Following Sec.~\ref{sec:local-content-supervised}, we map each block $B_j$ to its most relevant agenda item $z_j$:
\begin{equation}
\label{eq:cs-trajectory}
Z=\langle z_1,z_2,\ldots\rangle,\qquad 
z_j=\operatorname*{argmax}_{a\in\mathcal{A}} \sum_{u\in B_j}\text{sim}(e(u),e(a)).
\end{equation}
We then deduplicate adjacent repetitions in $Z$ to obtain
$\hat{Z}=\langle \hat{z}_1,\hat{z}_2,\ldots,\hat{z}_M\rangle$,
where $\hat{z}_{j}\neq \hat{z}_{j+1}$ for all $j$.

\textbf{Step 3 (Transitions orderliness relative to $\mathcal{L}$).}
We assess whether each consecutive transition $\hat{z}_j \rightarrow \hat{z}_{j+1}$ follows the reference order established in $\mathcal{L}$, allowing for local deviations. Intuitively, we consider a transition $\hat{z}_j \rightarrow \hat{z}_{j+1}$ to be \emph{in-order} if $\hat{z}_{j+1}$ is within a $r$-hop reach from $\hat{z}_j$ in the reference list $\mathcal{L}$ (in either direction); otherwise, we treat it as a \emph{regression}. Formally, let $\text{pos}(a)$ denote the index of agenda item $a$ in $\mathcal{L}$. We define regression as:
\begin{equation}
\label{eq:isreg}
\text{IsRegression}(a_i,a_j)
\;=\;
\mathbb{I}\!\left(\left|\text{pos}(a_j)-\text{pos}(a_i)\right|> r\right).
\end{equation}
i.e., $\text{IsRegression}(a_i,a_j)=0$ if $a_i \rightarrow a_j$ is in-order, and $\text{IsRegression}(a_i,a_j)=1$ otherwise. Therefore, CS is the fraction of non-regressive transitions in $\hat{Z}$:
\begin{equation}
\label{eq:cs}
\text{CS}(\hat{C},\mathcal{A})
\;=\;
1-\frac{1}{M-1}\sum_{j=1}^{M-1}\text{IsRegression}(\hat{z}_j,\hat{z}_{j+1}).
\end{equation}

A higher CS indicates a more disciplined, agenda-guided progression with few regressions (e.g., structured interviews or guided troubleshooting). A lower CS indicates frequent out-of-order transitions of agenda items, reflecting a more disjoint progression (e.g., uncontrolled topic hopping or repeated backtracking).

\subsubsection{Objective-free}
When no explicit objective or agenda is provided, global content quality is assessed by characterizing how the conversation evolves in semantic space. Concretely, we embed each utterance $\hat{u}_t$ in the generated conversation $\hat{C}$ as $e(\hat{u}_t)$, and quantify the overall progression pattern via embedding-space displacement and stepwise smoothness.

\paragraph{Progression Distance (PD)}
We define progression distance as the semantic displacement from the beginning to the end of $\hat{C}$, normalized by the number of generated turns. This measures how much the conversation's topic/content changes per turn on average:
\begin{equation}
\label{eq:pd}
\text{PD}(\hat{C}) \;=\; \frac{\left\|e(\hat{u}_{T}) - e(\hat{u}_{1})\right\|}{T},
\end{equation}
where $\|\cdot\|$ denotes Euclidean distance in the embedding space (and $T$ is the number of generated turns).

Lower PD suggests limited overall progression (e.g., redundant exchanges or stagnation). Higher PD indicates larger net semantic movement per turn; however, extremely high PD may also reflect abrupt topic drift rather than natural development, so PD is most informative when interpreted together with a smoothness measure such as HMP.

\paragraph{Harmonic Mean Progression (HMP)}
To capture steadier, incremental progression patterns, we compute the harmonic mean of consecutive turn-to-turn semantic step sizes:
\begin{equation}
\label{eq:hmp}
\text{HMP}(\hat{C})
\;=\;
\frac{T-1}
{\sum_{t=1}^{T-1}\frac{1}{\left\|e(\hat{u}_{t+1}) - e(\hat{u}_{t})\right\|+\epsilon}}.
\end{equation}
HMP penalizes trajectories with highly uneven step sizes (e.g., long stagnant discussions or occasional large jumps), resulting in a lower score; and favors conversations that maintain steady, relatively consistent semantic progression across turns.

\subsection{Global speaker--content Consistency}
\label{sec:global_speaker_content_consistency}

A direct baseline for evaluating global speaker--content consistency is to aggregate the local speaker--content consistency scores from Section~\ref{sec:local_speaker_content_consistency} over the entire conversation. Concretely, for each generated turn $(\hat{s}_t,\hat{u}_t)$ in $\hat{C}$, we compute a local speaker--content consistency score by comparing $\hat{u}_t$ against the set of prior messages attributed to the speaker $\hat{s}_t$, and then aggregate the per-turn scores across $\hat{C}$. Although this provides a natural extension of the local measure to the global setting, it incurs substantial computational overhead: for a conversation of length $n$, scoring turn $t$ may require comparing against up to $t$ prior messages, yielding $O(n^2)$ complexity in the worst case. For long-horizon generation, this quadratic cost can become prohibitive. We therefore introduce a more efficient centroid-based approach that captures global consistency while reducing the computation to linear complexity in conversation length.

Specifically, we adopt a prototype-based view of speaker behavior: a speaker's utterances in $\hat{C}$ can be modeled as one or more \emph{semantic clusters} (a.k.a., centroids), where each cluster corresponds to a characteristic of that speaker’s contributions (e.g., topical focus, functional role, or linguistic styles). Global speaker--content consistency is then measured by how well each utterance aligns with the speaker's cluster prototypes over the full interaction. Concretely, we represent each speaker using one or more centroid (i.e., semantic cluster center) embeddings and score utterances by their similarity to the closest centroid, yielding an efficient approximation that still penalizes long-horizon drift and inconsistent speaker behavior.

\paragraph{Single Centroid ($K=1$).}
For each speaker $s$, let $\hat{C}_s$ denote the set of utterances in $\hat{C}$ attributed to $s$. We embed each utterance $u\in \hat{C}_s$ as $e(u)$ and compute a single centroid that summarizes the speaker's overall contribution:
\begin{equation}
\label{eq:centroid_computation}
\mathbf{c}_{s} \;=\; \frac{1}{|\hat{C}_s|}\sum_{u\in \hat{C}_s} e(u).
\end{equation}
Intuitively, $\mathbf{c}_s$ serves as a prototype for the speaker's typical semantic content in $\hat{C}$. We then measure how well the speaker's utterances align with this prototype.

\textit{Average Distance to Centroid (GSCC-DC-avg).}
We measure the average semantic alignment between a speaker's utterances and the speaker centroid:
\begin{equation}
\label{eq:gscc-dc-avg}
\text{GSCC-DC-avg}(s) \;=\; \frac{1}{|\hat{C}_s|}\sum_{u\in \hat{C}_s} \text{sim}(e(u),\mathbf{c}_{s}).
\end{equation}

A high score indicates that the speaker's messages are tightly clustered around a consistent theme, suggesting strong topical coherence in their contributions. A low score indicates high variance, where the speaker's messages span diverse or disconnected topics.

\textit{Maximum Distance to Centroid (GSCC-DC-max).}
We also measure the strongest (best-case) alignment between any utterance and the speaker centroid:
\begin{equation}
\label{eq:gscc_dc_max_single}
\text{GSCC-DC-max}(s) \;=\; \max_{u \in \hat{C}_s} \text{sim} \bigl(e(u), \mathbf{c}_{s}\bigr).
\end{equation}

A higher GSCC-DC-max indicates that the speaker has at least one utterance that strongly matches their overall prototype; a lower value indicates that even the best-aligned utterance is only weakly consistent with the centroid, suggesting broad drift or unstable attribution. In practice, GSCC-DC-max complements GSCC-DC-avg by being less sensitive to occasional off-pattern turns.

\paragraph{Multiple Centroids ($K>1$).}
A single centroid implicitly assumes that a speaker's utterances are well described by a single semantic mode. In practice, a participant may exhibit multiple characteristics (e.g., discussing different subtopics or playing multiple functional roles). To capture such multifaceted semantic features, we represent each speaker $s$ with a set of centroids $\{\mathbf{c}_{s}^{(1)},\ldots,\mathbf{c}_{s}^{(K)}\}$ obtained by fitting a Gaussian Mixture Model (GMM) to the speaker's L2-normalized utterance embeddings. The number of centroids $K$ is selected automatically by minimizing the Bayesian Information Criterion (BIC), subject to an upper bound $K_{\max}=\left\lfloor \sqrt{|\hat{C}_s|} \right\rfloor$.

Given multiple centroids, each utterance is scored against its \emph{best-matching} centroid, allowing it to align with the most appropriate semantic characteristic. We define:
\begin{equation}
\label{eq:gscc_dc_avg_multi}
\text{GSCC-DC-avg}(s)
\;=\;
\frac{1}{|\hat{C}_s|}
\sum_{u \in \hat{C}_s}
\max_{j\in\{1,\ldots,K\}}
\text{sim} \bigl(e(u), \mathbf{c}_{s}^{(j)}\bigr),
\end{equation}
\begin{equation}
\label{eq:gscc_dc_max_multi}
\text{GSCC-DC-max}(s)
\;=\;
\max_{u \in \hat{C}_s}
\max_{j\in\{1,\ldots,K\}}
\text{sim} \bigl(e(u), \mathbf{c}_{s}^{(j)}\bigr).
\end{equation}
This formulation rewards messages that align with any of the speaker's semantic characteristic, while penalizing content that diverges from all characteristics.

\nop{
\paragraph{Incorporating Background Information} 
Similar to the local speaker--content consistency measures (Section~\ref{sec:local_speaker_content_consistency}), the centroid-based approach can be enhanced with speaker background information $B_{s_i}$ when available. We distinguish two cases:

\textbf{Case 1: Without background information.} The centroid is computed purely from message embeddings as defined in Equation~\ref{eq:centroid_computation}.

\textbf{Case 2: With background information.} When speakers have limited conversation history ($|\mathcal{C}_{s_i}|$ is small), we construct an augmented centroid that incorporates both message embeddings and background information. Let $e_{\text{message}}(s_i)$ denote the centroid computed from messages, and $e_{\text{background}}(s_i) = e(B_{s_i})$ denote the embedding of the speaker's profile. The augmented centroid is:
\begin{equation}
\mathbf{c}_{s_i}^{\text{aug}} = \alpha(s_i) \cdot e_{\text{message}}(s_i) + (1-\alpha(s_i)) \cdot e_{\text{background}}(s_i)
\end{equation}
where $\alpha(s_i) \in [0,1]$ is an adaptive weight based on the reliability of conversation history:
\begin{equation}
\alpha(s_i) = \min\left(1, \frac{|\mathcal{C}_{s_i}|}{k}\right)
\end{equation}
When the speaker has sufficient conversation history ($|\mathcal{C}_{s_i}| \geq k$), we rely entirely on observed messages ($\alpha = 1$); when history is sparse, we increasingly incorporate their documented expertise.
}

\paragraph{Incorporating Background Information.}
When available, speaker background information (e.g., a profile or role description) provides an additional reference for evaluating consistency, particularly when a speaker has only a few utterances in $\hat{C}$. We therefore augment the centroid-based formulation with background information, similar to Section~\ref{sec:local_speaker_content_consistency}.

Let $\mathbf{c}_s$ denote the centroid for speaker $s$ (Eq.~(\ref{eq:centroid_computation}), or its multi-centroid extension) derived from $\hat{C}_s$, and let $\mathbf{b}_s = e(B_s)$ denote the embedding of background information $B_s$. We define an augmented speaker prototype by a weighted combination of $\mathbf{c}_s$ and $\mathbf{b}_s$:
\begin{equation}
\label{eq:centroid_aug}
\mathbf{c}_s^{\text{aug}}
\;=\;
\alpha_s\,\mathbf{c}_s
\;+\;
\bigl(1-\alpha_s\bigr)\,\mathbf{b}_s,
\end{equation}
where $\alpha_s\in[0,1]$ is an adaptive weight that places more weight on $\mathbf{b}_s$ when $s$ has limited participation.

A higher background-aware global speaker--content consistency score indicates that a speaker's utterances align well with both their observed conversational behavior and their documented role/expertise. A lower score indicates mismatch with the speaker prototype---e.g., the speaker begins producing content inconsistent with their role or prior contributions---which may reflect long-horizon drift or speaker attribution errors. In particular, background augmentation is most informative when $|\hat{C}_s|$ is small.

\section{Detailed Generation Methods}
\label{sec:appendix_detailed_generation_methods}
We provide detailed descriptions and analysis of the multi-party conversation generation methods utilized in our benchmark experiments below. These methods are categorized based on their intended tasks: next-message prediction and full-conversation generation.

\subsection{Local Generation: Next-Message Prediction}
\label{subsec:appendix_local}

\textbf{{MultiLIGHT}~\cite{wei2023multi}:} This approach utilizes a two-stage fine-tuning paradigm that decouples the prediction of the next speaker from the generation of the utterance. Specifically, it employs a 400M-parameter BART model for speaker identification and a 3B-parameter model for content generation, conditioned on the conversation history and the predicted speaker. Both components are fine-tuned on a specialized dataset of multi-party fantasy role-playing dialogues using ground-truth speaker-utterance pairs. However, as a fine-tuning-based method, MultiLIGHT exhibits several limitations: it often reproduces training data patterns (limiting conversational diversity), demonstrates constrained generalizability across varying interaction dynamics or topics, and lacks multilingual support due to its English-only training corpus.

\textbf{{ChatGPT-solver}~\cite{tan2023chatgpt}:} This framework proposes a zero-shot prompting strategy for multi-party conversation generation by injecting explicit structural information into the prompt to aid \textsc{GPT-4} in interpreting multi-party dynamics. A significant limitation of this method is its inability to perform joint prediction; response generation requires the ground-truth next speaker as a prerequisite input, and conversely, speaker prediction requires predefined content.

\textbf{Standardized Prompt-Based LLMs:} Given the widespread adoption of prompt-based methods for multi-party conversation generation~\cite{wei2023multi, tan2023chatgpt}, we evaluate state-of-the-art (SOTA) LLM APIs, including \textsc{Llama-3.3-70B-Instruct}~\cite{dubey2024llama}, \textsc{GPT-4-Turbo}~\cite{achiam2023gpt}, \textsc{DeepSeek-Chat}~\cite{liu2024deepseek}, and \textsc{Claude-3.5-Sonnet}~\cite{Anthropic2024Claude35}. To address the absence of a model-agnostic standardized prompt format in existing literature, we utilize \textsc{Claude-3.5-Sonnet} to generate a unified prompt template as detailed in Appendix~\ref{app:prompt_prediction}. This ensures that comparisons across models are not biased by model-specific prompt engineering, facilitating a fair assessment of generation quality.

\subsection{Global Generation: Full-Conversation Generation}
\label{subsec:appendix_global}



\textbf{{MPC-constraints}~\cite{penzo2025don}:} This method provides a standardized prompt template designed for single-pass generation, where the LLM generates an entire multi-party conversation from scratch in a single execution given topic and speaker list. The template enforces constraints regarding output formatting, interaction dynamics, and speaker contribution guidelines. Although intended to be model-agnostic, the original study identified \textsc{Llama-3.1}~\cite{dubey2024llama} and \textsc{Qwen-2.5}~\cite{qwen25} as the most effective models for this task.

\textbf{{Extension to SOTA LLMs}:} Leveraging the model-agnostic framework provided by \textsc{MPC-constraints}, we extend the evaluation to a broader range of SOTA LLMs. In addition to the original top performers (\textsc{Llama-3.1} and \textsc{Qwen-2.5}), our benchmark includes \textsc{Llama-3.3-70B-Instruct}, \textsc{GPT-4-Turbo}, \textsc{DeepSeek-Chat}, and \textsc{Claude-3.5-Sonnet}.

\section{Additional Experimental Results}\label{sec:additional_experiment}
\subsection{Comparison of Generation Models: Local Evaluation}
Table~\ref{tab:local_speaker_measures_delidata}, Table~\ref{tab:local_content_measures_delidata}, Table~\ref{tab:local_speaker_content_consistency_delidata}, 
Table~\ref{tab:local_speaker_content_consistency_mpdd},
Table~\ref{tab:local_speaker_measures_mpdd}, and Table~\ref{tab:local_content_measures_mpdd} present additional experimental comparisons of different generation methods (including human-authored references) across the DeliData and MPDD datasets. These evaluations utilize \textsc{MPCEval}'s local metrics and are structured as follows:

\begin{itemize} \item \textbf{DeliData Experiments:} Table~\ref{tab:local_speaker_measures_delidata} reports local speaker modeling, Table~\ref{tab:local_content_measures_delidata} details local content quality, and Table~\ref{tab:local_speaker_content_consistency_delidata} covers local speaker--content consistency. \item \textbf{MPDD Experiments:} Table~\ref{tab:local_speaker_measures_mpdd} presents results for local speaker modeling, Table~\ref{tab:local_content_measures_mpdd} for local content quality, and Table~\ref{tab:local_speaker_content_consistency_mpdd} for local speaker--content consistency. \end{itemize}

Note that the MPDD dataset is inapplicable to \textsc{MultiLIGHT}, as the latter is restricted to English text generation. Furthermore, local speaker--content consistency metrics cannot evaluate \textsc{ChatGPT-solver} due to the method's architectural requirement for ground-truth speakers during content generation and vice versa (see Appendix~\ref{subsec:appendix_local}).

\begin{table*}[t]
\centering
\caption{Performance comparison on the Delidata dataset using \textsc{MPCEval}'s local speaker modeling measures (mean $\pm$ standard deviation). \textbf{Bold} and \underline{underlined} values denote the highest and lowest scores in each column, respectively.}
\label{tab:local_speaker_measures_delidata}
\resizebox{\textwidth}{!}{%
\begin{tabular}{lcccccc}
\toprule
& \multicolumn{6}{c}{\textbf{Local Speaker Modeling}} \\
\cmidrule(lr){2-7} 
\textbf{Model} & \textbf{DNR} & \textbf{IR} & \textbf{PF} & \textbf{LS-ES-avg} & \textbf{LS-ES-max} & \textbf{LS-TA} \\
\midrule
\textsc{LLaMa-3.3} & 0.278 $\pm$ 0.448 & 0.182 $\pm$ 0.205 & \underline{0.276 $\pm$ 0.096} & \underline{0.444 $\pm$ 0.088} & 0.845 $\pm$ 0.154 & 0.778 $\pm$ 0.193 \\
\textsc{GPT-4-Turbo} & 0.278 $\pm$ 0.448 & 0.185 $\pm$ 0.203 & 0.303 $\pm$ 0.100 & 0.456 $\pm$ 0.097 & 0.838 $\pm$ 0.149 & 0.783 $\pm$ 0.193 \\
\textsc{Deepseek} & \textbf{0.389 $\pm$ 0.488} & 0.213 $\pm$ 0.223 & 0.278 $\pm$ 0.099 & 0.450 $\pm$ 0.100 & 0.831 $\pm$ 0.158 & \underline{0.774 $\pm$ 0.196} \\
\textsc{Claude-3.5} & 0.333 $\pm$ 0.471 & \underline{0.125 $\pm$ 0.143} & 0.298 $\pm$ 0.101 & 0.454 $\pm$ 0.096 & 0.837 $\pm$ 0.149 & 0.836 $\pm$ 0.202 \\
\midrule
\textsc{MultiLIGHT}   &  0.333 $\pm$ 0.471 & 0.227 $\pm$ 0.216 & 0.309 $\pm$ 0.102 & 0.448 $\pm$ 0.088 & \underline{0.819 $\pm$ 0.145} & 0.781 $\pm$ 0.174 \\
\textsc{ChatGPT-solver}  & 0.333 $\pm$ 0.471 & 0.195 $\pm$ 0.196 & \textbf{0.369 $\pm$ 0.126} & \textbf{0.489 $\pm$ 0.098} & \textbf{0.866 $\pm$ 0.125} & \textbf{0.848 $\pm$ 0.160} \\
\midrule
Human & \underline{0.222 $\pm$ 0.416} & \textbf{0.295 $\pm$ 0.254} & 0.312 $\pm$ 0.110 & 0.458 $\pm$ 0.103 & 0.825 $\pm$ 0.160 & 0.816 $\pm$ 0.191 \\ 
\bottomrule
\end{tabular}%
}
\end{table*}

\begin{table*}[t]
\centering
\caption{Performance comparison on the DeliData dataset using \textsc{MPCEval}'s local content quality measures (mean $\pm$ standard deviation). \textbf{Bold} and \underline{underlined} values denote the highest and lowest scores in each column, respectively. For \textsc{LNR-E-w}, M-SNS-min, M-SNS-avg, and \textsc{TES}, values closest to 0.5 are \textbf{bolded}, while those farthest from 0.5 are \underline{underlined}.}
\label{tab:local_content_measures_delidata}
\resizebox{\textwidth}{!}{%
\begin{tabular}{lcccccc}
\toprule
& \multicolumn{6}{c}{\textbf{Local Content Quality}} \\
\cmidrule(lr){2-7} 
\textbf{Model} & \textbf{LNR-E-w} & \textbf{M-SNS-min} & \textbf{M-SNS-avg} & \textbf{DAF} & \textbf{LL} & \textbf{TES} \\
\midrule
\textsc{LLaMa-3.3} & $0.280 \pm 0.159$ & $0.336 \pm 0.095$ & $0.735 \pm 0.091$ & $0.281 \pm 0.243$ & $0.893 \pm 0.091$ & $0.357 \pm 0.399$ \\
\textsc{GPT-4-Turbo}  & $\textbf{0.403} \pm 0.177$ & $\textbf{0.376} \pm 0.131$ & $0.736 \pm 0.077$ & $\textbf{0.363} \pm 0.237$ & $0.793 \pm 0.178$ & $0.142 \pm 0.168$ \\
\textsc{Deepseek} & $0.314 \pm 0.133$ & $0.283 \pm 0.095$ & $\textbf{0.688} \pm 0.061$ & $0.307 \pm 0.240$ & $\textbf{0.918} \pm 0.074$ & \underline{$0.112 \pm 0.185$}\\
\textsc{Claude-3.5} & $0.343 \pm 0.176$ & $0.332 \pm 0.125$ & $0.698 \pm 0.081$ & \underline{$0.279 \pm 0.236$} & $0.838 \pm 0.203$ & $\textbf{0.500} \pm 0.497$\\
\midrule
\textsc{MultiLIGHT}   & \underline{0.156 $\pm$ 0.233} & \underline{0.233 $\pm$ 0.154} & \underline{0.749 $\pm$ 0.072} & $0.329 \pm 0.267$ & $0.757 \pm 0.222$ & $0.303 \pm 0.407$ \\ 
\textsc{ChatGPT-solver} & $0.289 \pm 0.202$ & $0.298 \pm 0.113$ & $0.690 \pm 0.068$ & $0.294 \pm 0.238$ & $0.833 \pm 0.192$ & $0.167 \pm 0.373$  \\
\midrule
Human & $0.245 \pm 0.299$ & $0.250 \pm 0.233$ & $0.719 \pm 0.109$ & $0.348 \pm 0.234$ & \underline{0.232 $\pm$ 0.349} & $0.281 \pm 0.383$\\
\bottomrule
\end{tabular}%
}
\end{table*}

\begin{table*}[t]
\centering
\begin{minipage}{0.48\textwidth}
    \centering
    \caption{Performance comparison on the Delidata dataset using \textsc{MPCEval}'s local speaker--content consistency measures (mean $\pm$ standard deviation). \textbf{Bold} and \underline{underlined} values denote the highest and lowest scores in each column, respectively.}
    \label{tab:local_speaker_content_consistency_delidata}
    \small
    \begin{tabular}{lccc}
    \toprule
    & \multicolumn{3}{c}{\textbf{Local Speaker--Content Consistency}} \\
    \cmidrule(lr){2-4} 
    \textbf{Model} & \textbf{LSCC-ES-avg} & \textbf{LSCC-ES-max} & \textbf{LSCC-ES-min} \\
    \midrule
    \textsc{LLaMa-3.3} & $0.358 \pm 0.198$ & $0.388 \pm 0.223$ & $0.328 \pm 0.195$ \\
    \textsc{GPT-4-Turbo} & $\mathbf{0.421 \pm 0.206}$ & $\mathbf{0.446 \pm 0.209}$ & $\mathbf{0.396 \pm 0.220}$ \\
    \textsc{Deepseek}  & $0.381 \pm 0.161$ & $0.413 \pm 0.167$ & $0.349 \pm 0.173$ \\
    \textsc{Claude-3.5} & $0.331 \pm 0.178$ & \underline{$0.356 \pm 0.199$} & $0.306 \pm 0.174$ \\
    \midrule
    \textsc{MultiLIGHT}    &  \underline{$0.297 \pm 0.211$} & $0.370 \pm 0.240$ & \underline{$0.232 \pm 0.219$} \\  
    \textsc{ChatGPT-solver} & - &- & -  \\
    \midrule
    Human & $0.334 \pm 0.203$ & $0.388 \pm 0.198$ & $0.280 \pm 0.220$ \\  
    \bottomrule
    \end{tabular}
\end{minipage}
\hfill
\begin{minipage}{0.48\textwidth}
    \centering
    \caption{Performance comparison on the MPDD dataset using \textsc{MPCEval}'s local speaker--content consistency measures (mean $\pm$ standard deviation). \textbf{Bold} and \underline{underlined} values denote the highest and lowest scores in each column, respectively.}
    \label{tab:local_speaker_content_consistency_mpdd}
    \small
    \begin{tabular}{lccc}
    \toprule
    & \multicolumn{3}{c}{\textbf{Local Speaker--Content Consistency}} \\
    \cmidrule(lr){2-4} 
    \textbf{Model} & \textbf{LSCC-ES-avg} & \textbf{LSCC-ES-max} & \textbf{LSCC-ES-min} \\
    \midrule
    \textsc{LLaMa-3.3}    & $\mathbf{0.452 \pm 0.139}$ & $\mathbf{0.536 \pm 0.161}$ & $0.370 \pm 0.156$ \\
    \textsc{GPT-4-Turbo} & \underline{$0.400 \pm 0.139$} & \underline{$0.475 \pm 0.156$} & \underline{$0.326 \pm 0.155$}  \\
    \textsc{Deepseek} & 0.450 $\pm$ 0.142 & 0.527 $\pm$ 0.158 & \textbf{0.375 $\pm$ 0.161} \\
    \textsc{Claude-3.5}   &  $0.433 \pm 0.140$  &  $0.506 \pm 0.157$  &   $0.360 \pm 0.155$    \\
    \midrule
    MultiLIGHT & - & - & - \\
    ChatGPT-solver &  -  &  -  &   -    \\
    \midrule
    Human & $0.417 \pm 0.138$ & $0.488 \pm 0.158$ & $0.345 \pm 0.149$ \\
    \bottomrule
    \end{tabular}
\end{minipage}
\end{table*}

\nop{
\begin{table*}[t]
\centering
\caption{Performance comparison of different LLMs on the DeliData dataset using the proposed local speaker--content consistency measures. Results are reported as mean $\pm$ standard deviation. The best performance is highlighted in bold.}
\label{tab:speaker_content_consistency_delidata}
\begin{tabular}{lccc}
\toprule
& \multicolumn{3}{c}{\textbf{Speaker Content Consistency Measures}} \\
\cmidrule(lr){2-4} 
\textbf{Model} & \textbf{LSCC-ES-avg} & \textbf{LSCC-ES-max} & \textbf{LSCC-ES-min} \\
\midrule
MultiLIGHT   &  $0.297 \pm 0.211$ & $0.370 \pm 0.240$ & $0.232 \pm 0.219$ \\ 
ChatGPT-solver & - &- & -  \\
\midrule
llama-3.3-70b-instruct & $0.358 \pm 0.198$ & $0.388 \pm 0.223$ & $0.328 \pm 0.195$ \\
gpt-4-turbo & $0.421 \pm 0.206$ & $0.446 \pm 0.209$ & $0.396 \pm 0.220$ \\
deepseek-chat  & $0.381 \pm 0.161$ & $0.413 \pm 0.167$ & $0.349 \pm 0.173$ \\
claude-3.5-sonnet & $0.331 \pm 0.178$ & $0.356 \pm 0.199$ & $0.306 \pm 0.174$ \\
\midrule
\textbf{Ground-truth} & $0.334 \pm 0.203$ & $0.388 \pm 0.198$ & $0.280 \pm 0.220$ \\ 
\bottomrule
\end{tabular}
\end{table*}
}

\begin{table*}[t]
\centering
\caption{Performance comparison on the MPDD dataset using \textsc{MPCEval}'s local speaker modeling measures (mean $\pm$ standard deviation). \textbf{Bold} and \underline{underlined} values denote the highest and lowest scores in each column, respectively. }
\label{tab:local_speaker_measures_mpdd}
\resizebox{\textwidth}{!}{%
\begin{tabular}{lcccccc}
\toprule
& \multicolumn{6}{c}{\textbf{Local Speaker Modeling}} \\
\cmidrule(lr){2-7} 
\textbf{Model} & \textbf{DNR} & \textbf{IR} & \textbf{PF} & \textbf{LS-ES-avg} & \textbf{LS-ES-max} & \textbf{LS-TA} \\
\midrule
\textsc{LLaMa-3.3}   &  0.102 $\pm$ 0.302 & \underline{0.440 $\pm$ 0.222} & \underline{0.400 $\pm$ 0.134} & 0.688 $\pm$ 0.117 & 0.789 $\pm$ 0.142 & 0.660 $\pm$ 0.119 \\ 
\textsc{GPT-4-Turbo}  & 0.100 $\pm$ 0.301 & 0.468 $\pm$ 0.210 & 0.408 $\pm$ 0.127 & 0.692 $\pm$ 0.109 & 0.793 $\pm$ 0.134 & 0.666 $\pm$ 0.112 \\
\textsc{Deepseek} & \textbf{0.105 $\pm$ 0.306} & 0.463 $\pm$ 0.211 & \underline{0.400 $\pm$ 0.131} & 0.689 $\pm$ 0.122 & 0.792 $\pm$ 0.136 & 0.661 $\pm$ 0.115 \\
\textsc{Claude-3.5}    &  $0.102 \pm 0.303$ & $0.474 \pm 0.212$ & $0.401 \pm 0.133$ & $0.689 \pm 0.114$ & $0.791 \pm 0.138$ & $0.660 \pm 0.117$ \\
\midrule
\textsc{MultiLIGHT}   &  - & - & - & - & - & - \\ 
\textsc{ChatGPT-solver} & \underline{$0.089 \pm 0.284$} & $0.462 \pm 0.208$ & \textbf{0.421 $\pm$ 0.129} & \textbf{0.702 $\pm$ 0.077} & \textbf{0.803 $\pm$ 0.099} & \textbf{0.677 $\pm$ 0.084} \\
\midrule
Human & $0.094 \pm 0.292$ & $\mathbf{0.498 \pm 0.200}$ & \underline{$0.400 \pm 0.139$} & \underline{$0.677 \pm 0.149$} & \underline{$0.775 \pm 0.176$} & \underline{$0.649 \pm 0.149$} \\
\bottomrule
\end{tabular}%
}
\end{table*}

\begin{table*}[t]
\centering
\caption{Performance comparison on the MPDD dataset using \textsc{MPCEval}'s local content quality measures (mean $\pm$ standard deviation). \textbf{Bold} and \underline{underlined} values denote the highest and lowest scores in each column, respectively; for \textsc{LNR-E-w}, M-SNS-min, M-SNS-avg, and \textsc{TES}, values closest to 0.5 are \textbf{bolded}, while those farthest from 0.5 are \underline{underlined}.}
\label{tab:local_content_measures_mpdd}
\resizebox{\textwidth}{!}{%
\begin{tabular}{lcccccc}
\toprule
& \multicolumn{6}{c}{\textbf{Local Content Quality}} \\
\cmidrule(lr){2-7} 
\textbf{Model} & \textbf{LNR-E-w} & \textbf{M-SNS-min} & \textbf{M-SNS-avg} & \textbf{DAF} & \textbf{LL} & \textbf{TES} \\
\midrule
\textsc{LLaMa-3.3}   &   \textbf{0.545 $\pm$ 0.200} & $0.341 \pm 0.123$ & $0.577 \pm 0.094$ & \underline{$0.405 \pm 0.290$} & $0.641 \pm 0.234$ & $0.450 \pm 0.404$   \\
\textsc{GPT-4-Turbo}  &  $0.588 \pm 0.179$ & $\mathbf{0.376 \pm 0.127}$ & \underline{$0.594 \pm 0.100$} & $0.427 \pm 0.285$ & $0.678 \pm 0.203$ & $0.447 \pm 0.407$ \\
\textsc{Deepseek} & 0.572 $\pm$ 0.177 & 0.341 $\pm$ 0.118 & 0.562 $\pm$ 0.096 & \textbf{0.428 $\pm$ 0.283} & \underline{0.825 $\pm$ 0.151} & \textbf{0.453 $\pm$ 0.406} \\
\textsc{Claude-3.5} &  \underline{$0.589 \pm 0.164$} & $0.359 \pm 0.114$ & $0.573 \pm 0.095$ & $0.426 \pm 0.286$ & $0.788 \pm 0.165$ & $0.446 \pm 0.407$\\
\midrule
\textsc{MultiLIGHT} & - & - & - & - & - & - \\
\textsc{ChatGPT-solver} &  $0.579 \pm 0.180$ & \underline{$0.333 \pm 0.117$} & \textbf{0.559 $\pm$ 0.096} & $0.426 \pm 0.283$ & $0.774 \pm 0.202$ & \underline{$0.437 \pm 0.404$}\\
\midrule
\textbf{Ground-truth} & $0.586 \pm 0.274$ & $0.366 \pm 0.140$ & $0.586 \pm 0.098$ & $0.416 \pm 0.286$ & \underline{$0.464 \pm 0.291$} & $0.443 \pm 0.408$\\
\bottomrule
\end{tabular}%
}
\end{table*}

\nop{
\begin{table*}[t]
\centering
\caption{Performance comparison of different LLMs on the MPDD dataset (1,774 dialogues) using the proposed local speaker--content consistency measures. Results are reported as mean $\pm$ standard deviation. The best performance is highlighted in bold.}
\label{tab:speaker_content_consistency_mpdd}
\begin{tabular}{lccc}
\toprule
& \multicolumn{3}{c}{\textbf{Speaker Content Consistency Measures}} \\
\cmidrule(lr){2-4} 
\textbf{Model} & \textbf{LSCC-ES-avg} & \textbf{LSCC-ES-max} & \textbf{LSCC-ES-min} \\
\midrule
llama-3.3-70b-instruct   & $\mathbf{0.452 \pm 0.139}$ & $\mathbf{0.536 \pm 0.161}$ & $0.370 \pm 0.156$ \\
gpt-4-turbo & $0.400 \pm 0.139$ & $0.475 \pm 0.156$ & $0.326 \pm 0.155$  \\
deepseek-chat & 0.450 $\pm$ 0.142 & 0.527 $\pm$ 0.158 & \textbf{0.375 $\pm$ 0.161} \\
claude-3.5-sonnet  &  $0.433 \pm 0.140$  &  $0.506 \pm 0.157$  &   $0.360 \pm 0.155$    \\
\midrule
MultiLIGHT & - & - & - \\
ChatGPT-solver &  -  &  -  &   -    \\
\midrule
\textbf{Ground-truth} & $0.417 \pm 0.138$ & $0.488 \pm 0.158$ & $0.345 \pm 0.149$ \\
\bottomrule
\end{tabular}
\end{table*}
}

\subsection{Comparison of Generation Models: Global Evaluation}
Table~\ref{tab:global_delidata} and Table~\ref{tab:global_tanka} present comparative experiments of various generation methods (including human-authored references) on the DeliData and Tanka datasets, respectively. Each table provides a global evaluation across three dimensions: global speaker modeling, global content quality, and global speaker--content consistency.

\begin{table*}[t]
\centering
\caption{Performance comparison on the Delidata dataset using \textsc{MPCEval}'s global measures, including global speaker modeling, global content quality, and global speaker--content consistency (mean $\pm$ standard deviation). \textbf{Bold} and \underline{underlined} values denote the highest and lowest scores in each column, respectively.}
\label{tab:global_delidata}
\resizebox{\textwidth}{!}{%
\begin{tabular}{lcc|cc|cccc}
\toprule
& \multicolumn{2}{c|}{\textbf{Global Speaker Modeling}} 
& \multicolumn{2}{c|}{\textbf{Global Content Quality}} 
& \multicolumn{4}{c}{\textbf{Global Speaker--Content Consistency}} \\
\cmidrule(lr){2-3}\cmidrule(lr){4-5}\cmidrule(lr){6-9}
& & & & & \multicolumn{2}{c|}{\textbf{Single Centroid ($k=1$)}} & \multicolumn{2}{c}{\textbf{Multiple Centroids ($k>1$)}} \\
\cmidrule(lr){6-7}\cmidrule(lr){8-9}
\textbf{Model} & \textbf{NSE} & \textbf{SC-Gini} & \textbf{PD} & \textbf{HMP} & \textbf{GSCC-DC-avg} & \textbf{GSCC-DC-max} & \textbf{GSCC-DC-avg} & \textbf{GSCC-DC-max} \\
\midrule
\textsc{MPC-constrains + LLaMa-3.1} & $0.978 \pm 0.022$ & $\mathbf{0.445 \pm 0.197}$ & $1.127 \pm 0.163$ & $0.869 \pm 0.107$ & $0.894 \pm 0.040$ & $0.898 \pm 0.040$ & $0.894 \pm 0.040$ & $0.898 \pm 0.040$ \\
\textsc{MPC-constrains + Qwen} & $0.983 \pm 0.024$ & $0.392 \pm 0.176$ & $\mathbf{1.648 \pm 0.215}$ & $\mathbf{1.021 \pm 0.114}$ & $\mathbf{0.951 \pm 0.061}$ & $\mathbf{0.951 \pm 0.061}$ & $\mathbf{0.951 \pm 0.061}$ & $\mathbf{0.951 \pm 0.061}$ \\
\midrule
\textsc{MPC-constrains + LLaMa-3.3} & \underline{$0.960 \pm 0.000$} & $0.430 \pm 0.163$ & $1.144 \pm 0.148$ & $0.835 \pm 0.074$ & $0.915 \pm 0.026$ & $0.915 \pm 0.026$ & $0.915 \pm 0.026$ & $0.915 \pm 0.026$ \\
\textsc{MPC-constrains + GPT-4-Turbo} & $0.984 \pm 0.020$ & $0.420 \pm 0.185$ & $1.198 \pm 0.232$ & $0.875 \pm 0.095$ & $0.885 \pm 0.039$ & $0.888 \pm 0.035$ & $0.885 \pm 0.039$ & $0.888 \pm 0.035$ \\
\textsc{MPC-constrains + Deepseek} & $0.979 \pm 0.024$ & $0.416 \pm 0.197$ & $1.270 \pm 0.145$ & \underline{$0.807 \pm 0.095$} & $0.918 \pm 0.041$ & $0.918 \pm 0.041$ & $0.918 \pm 0.041$ & $0.918 \pm 0.041$ \\
\textsc{MPC-constrains + Claude-3.5} & $\mathbf{0.987 \pm 0.015}$ & $0.366 \pm 0.203$ & $1.280 \pm 0.115$ & $0.967 \pm 0.070$ & $0.862 \pm 0.036$ & $0.867 \pm 0.033$ & $0.862 \pm 0.036$ & $0.867 \pm 0.033$ \\
\midrule
Human & $0.962 \pm 0.024$ & \underline{$0.245 \pm 0.182$} & \underline{$0.850 \pm 0.198$} & $0.873 \pm 0.484$ & \underline{$0.528 \pm 0.065$} & \underline{$0.667 \pm 0.045$} & \underline{$0.543 \pm 0.061$} & \underline{$0.691 \pm 0.055$} \\
\bottomrule
\end{tabular}
}
\end{table*}

\begin{table*}[t]
\centering
\caption{Performance comparison on the Tanka dataset using \textsc{MPCEval}'s global measures, including global speaker modeling, global content quality, and global speaker--content consistency (mean $\pm$ standard deviation). \textbf{Bold} and \underline{underlined} values denote the highest and lowest scores in each column, respectively.}
\label{tab:global_tanka}
\resizebox{\textwidth}{!}{%
\begin{tabular}{lcc|cc|cccc}
\toprule
& \multicolumn{2}{c|}{\textbf{Global Speaker Modeling}} 
& \multicolumn{2}{c|}{\textbf{Global Content Quality}} 
& \multicolumn{4}{c}{\textbf{Global Speaker--Content Consistency}} \\
\cmidrule(lr){2-3}\cmidrule(lr){4-5}\cmidrule(lr){6-9}
& & & & & \multicolumn{2}{c|}{\textbf{Single Centroid ($k=1$)}} & \multicolumn{2}{c}{\textbf{Multiple Centroids ($k>1$)}} \\
\cmidrule(lr){6-7}\cmidrule(lr){8-9}
\textbf{Model} & \textbf{NSE} & \textbf{SC-Gini} & \textbf{PD} & \textbf{HMP} & \textbf{GSCC-DC-avg} & \textbf{GSCC-DC-max} & \textbf{GSCC-DC-avg} & \textbf{GSCC-DC-max} \\
\midrule
\textsc{LLaMa-3.3} & $0.796 \pm 0.082$ & $0.573 \pm 0.118$ & $0.559 \pm 0.122$ & $0.620 \pm 0.319$ & $\mathbf{0.841 \pm 0.048}$ & $0.899 \pm 0.039$ & $0.869 \pm 0.057$ & $0.933 \pm 0.049$ \\
\textsc{GPT-4-Turbo} & $0.801 \pm 0.103$ & $0.557 \pm 0.140$ & $0.516 \pm 0.094$ & $\underline{0.592 \pm 0.252}$ & $0.832 \pm 0.025$ & $0.893 \pm 0.013$ & $0.834 \pm 0.025$ & $0.895 \pm 0.013$ \\
\textsc{Deepseek} & $0.770 \pm 0.067$ & $0.548 \pm 0.107$ & $\underline{0.466 \pm 0.088}$ & $0.738 \pm 0.249$ & $0.837 \pm 0.027$ & $\mathbf{0.903 \pm 0.022}$ & $\mathbf{0.909 \pm 0.045}$ & $\mathbf{0.974 \pm 0.025}$ \\
\textsc{Claude-3.5} & $\mathbf{0.822 \pm 0.078}$ & $\underline{0.543 \pm 0.173}$ & $\mathbf{0.572 \pm 0.084}$ & $0.725 \pm 0.289$ & $0.797 \pm 0.019$ & $0.873 \pm 0.021$ & $0.809 \pm 0.017$ & $0.892 \pm 0.019$ \\
\midrule
Human & $\underline{0.691 \pm 0.068}$ & $\mathbf{0.670 \pm 0.082}$ & $0.552 \pm 0.111$ & $\mathbf{0.798 \pm 0.238}$ & $\underline{0.723 \pm 0.031}$ & \underline{$0.835 \pm 0.017$} & \underline{$0.752 \pm 0.024$} & \underline{$0.879 \pm 0.039$} \\
\bottomrule
\end{tabular}
}
\end{table*}

\subsection{Case Study: Human-Authored Response is not the Sole Gold Standard}
\begin{figure*}[t]
\centering
\small

\setlength{\tabcolsep}{4pt}

\fcolorbox{black}{gray!10}{%
  \parbox{\dimexpr\textwidth-2\fboxsep-2\fboxrule\relax}{%
    \scriptsize
    \textbf{Context:} This conversation is from the DeliData dataset, where participants collaboratively solve the Wason card selection task. In this specific instance, four cards are presented showing: \textbf{3}, \textbf{2}, \textbf{I}, and \textbf{R}. Each card has a letter on one side and a number on the other. Participants must determine which cards to flip to test the rule: ``If a card has a vowel on one side, it has an even number on the other side.'' The participants are debating which cards to select to verify or falsify this rule.
  }%
}
\vspace{4pt}

\begin{tabularx}{\textwidth}{|>{\raggedright\arraybackslash}X|>{\raggedright\arraybackslash}p{0.38\textwidth}|}
\hline
\multicolumn{1}{|c|}{\textbf{Conversation History (Last 10 Turns)}} &
\multicolumn{1}{c|}{\textbf{Next Message \& Metrics}} \\
\hline

\scriptsize
\textbf{Leopard:} but it would make sense to falsify the rule by flipping 3 \newline
\textbf{Bat:} but the rule is say with the vowel card there is even number on the other side \newline
\textbf{Leopard:} as if 3 had a vowel it would be incorrect \newline
\textbf{Lion:} Yes but I am not fussed if it's incorrect if I can't see it, I'm not looking to disprove the rule I'm looking to prove \newline
\textbf{Lion:} Thats my thinking \newline
\textbf{Leopard:} right.... \newline
\textbf{Lion:} FINAL ANSWER 2 \newline
\textbf{Leopard:} what about i \newline
\textbf{Lion:} That's a vowel that can stay \newline
\textbf{Leopard:} are they gonna give us the correct answer \newline
\textbf{Lion:} I'm looking for vowels, I hope not \newline
\textbf{Leopard:} can you give us the correct answer when we are done \newline
\textbf{Leopard:} thanks!
&
\scriptsize
\vspace{-4pt}
\fbox{%
  \parbox{\dimexpr\linewidth-2\fboxsep-2\fboxrule\relax}{%
    \textbf{LLM-Generated Next Message:}\newline
    \textbf{Leopard:} Sure, let's finish up and then we'll see if we got it right!
  }%
}
\vspace{2pt}

\fbox{%
  \parbox{\dimexpr\linewidth-2\fboxsep-2\fboxrule\relax}{%
    \textbf{Human-Authored Next Message:}\newline
    \textbf{Leopard:} @all where you gone
  }%
}
\vspace{2pt}

\textbf{Metric Comparison:}

\vspace{1pt}
\renewcommand{\arraystretch}{1.2}
\begin{tabular}{@{}lcc@{}}
& \textbf{LLM} & \textbf{Human} \\
\hline
LNR-E-w   & \textbf{0.475} & 0.753 \\
M-SNS-avg & \textbf{0.714} & 0.825 \\
LL        & \textbf{0.924} & 0.038 \\
\end{tabular}
\\

\hline
\end{tabularx}

\caption{Case study. Human-authored next message is not the only valid response.}
\label{fig:case-study-rq3}
\end{figure*}
As indicated in Figure~\ref{fig:case-study-rq3}, the conversation progresses through a collaborative problem-solving on card selection. After Lion declares ``FINAL ANSWER 2,'' Leopard requests confirmation (``can you give us the correct answer when we are done'') and acknowledges with ``thanks!''. At this juncture, the human-authored next message shows Leopard expressing confusion or disorientation (``@all where you gone''), suggesting the speaker has lost track of the discussion flow. In contrast, the model generates a next message that provides clarity and direction (``Sure, let's finish up and then we'll see if we got it right!''), directly addressing the earlier request while proposing task completion.

The local content quality measures show clear differences between these two responses. The generated next message achieves significantly higher log-likelihood (LL) (0.924 vs. 0.038), reflecting its stronger alignment with probable continuations given the immediate context. Conversely, the human-authored next message's substantially lower LL (0.038) captures its unexpected expressions of confusion. The weighted embedding-aware lexical novelty (LNR-E-w) further distinguishes the two responses: the generated next message shows moderate novelty (0.475), close to the optimal range of $\approx$ 0.5, indicating it stays on topic while adding new information (i.e., giving direction for next step). The human-authored next message shows higher novelty (0.753), indicating it introduces concepts (i.e., confusion) that diverge from the recent discussion. Similarly, semantic novelty metrics (M-SNS-avg) distinguish between the generated next message's progression (0.714) and the human-authored next message's greater semantic deviation (0.825).

This case illustrates how our proposed metrics in \textsc{MPCEval} effectively capture the distinct characteristics between responses. The human-authored next message reflects a moment of confusion during collaborative problem-solving, while the generated next message maintains direct task engagement.

Moreover, this case study demonstrates that multiple valid continuations can coexist in complex multi-party dialogues, and a human-authored response should not be viewed as the sole ``gold standard'' reference. Both responses analyzed here are contextually plausible; however, they pursue different conversational trajectories. Consequently, if one relies strictly on traditional reference-based metrics that penalize any deviation from a single human-authored reference, a model-generated response, despite being a high-quality and coherent continuation, would receive a deceptively low score. This highlights the necessity of reference-free evaluation suites like \textsc{MPCEval}, which assess intrinsic qualities such as task alignment and local coherence rather than simple lexical overlap with a single reference.

\section{Prompt for Next Message Prediction}
\label{app:prompt_prediction}

The template used for batch next message prediction is presented in Box~\ref{box:prediction_prompt}.

\begin{figure*}[ht]
\begin{promptbox}[label={box:prediction_prompt}]{Next Message Prediction Template}
\small
\noindent I will give you multiple dialogues. For each dialogue, predict both the most likely next speaker AND the most likely next message content that speaker would say. Make sure the speaker and message content are consistent with each other.

\vspace{3mm}
\noindent Respond with ONLY a JSON array containing the dialogue\_id, predicted next\_speaker, and predicted next\_message for each dialogue, like this:
\begin{center}
\begin{minipage}{0.95\textwidth}
\begin{verbatim}
[
  {"dialogue_id": "1", "next_speaker": "SpeakerName", 
   "next_message": "The message content that speaker would say"},
  {"dialogue_id": "2", "next_speaker": "AnotherSpeaker", 
   "next_message": "Another message content"}
]
\end{verbatim}
\end{minipage}
\end{center}

\vspace{2mm}
\noindent Here are the dialogues:

\vspace{2mm}
\noindent \texttt{\dots} \\
\texttt{--- Dialogue \{idx\} (ID: \{dialogue['dialogue\_id']\}) ---} \\
\texttt{\{turn['speaker']\}: \{turn['utterance']\}} \\
\texttt{\dots}

\vspace{4mm}
\noindent Now respond with the JSON array of predictions (both next\_speaker and next\_message for each dialogue):
\end{promptbox}
\end{figure*}


\end{sloppy}
\end{document}